# Web-based Argumentation

Kenrick <kenr0002@e.ntu.edu.sg>
Technical Report
School of Computer Science and Engineering
Nanyang Technological University

## Abstract

Assumption-Based Argumentation (ABA) is an argumentation framework that has been proposed in the late 20th century. Since then, there was still no solver implemented in a programming language which is easy to setup and no solver have been interfaced to the web, which impedes the interests of the public. This project aims to implement an ABA solver in a modern programming language that performs reasonably well and interface it to the web for easier access by the public. This project has demonstrated the novelty of development of an ABA solver, that computes *conflict-free*, *stable*, *admissible*, *grounded*, *ideal*, and *complete* semantics, in Python programming language which can be used via an easy-to-use web interface for visualization of the argument and dispute trees. Experiments were conducted to determine the project's best configurations and to compare this project with *proxdd*, a state-of-the-art ABA solver, which has no web interface and computes less number of semantics. From the results of the experiments, this project's best configuration is achieved by utilizing "pickle" technique and tree caching technique. Using this project's best configuration, this project achieved a lower average runtime compared to *proxdd*. On other aspect, this project encountered more cases with exceptions compared to *proxdd*, which might be caused by this project computing more semantics and hence requires more resources to do so. Hence, it can be said that this project run comparably well to the state-of-the-art ABA solver *proxdd*. Future works of this project include computational complexity analysis and efficiency analysis of algorithms implemented, implementation of more semantics in argumentation framework, and usability testing of the web interface.

# 1. Introduction

One of the goals of Artificial Intelligence (AI) field is to achieve an intelligent machine. A machine can be considered as intelligent when it behaves like a human. One of the most pervasive human activities that are distinct from the other species is the ability to argue. Arguing, informally speaking, is a form of interaction among the participants to give rationales for agreeing or disagreeing with claims. This is a universal human ability and can be seen everywhere in daily human activities. Hence, it led to many studies regarding human ability to argue.

Within artificial intelligence, the field that is closely related to argumentation [1] is Knowledge Representation (KR). This field aims to represent information about the world into a computable form to solve complicated tasks. Solving these tasks require a complicated logic to automate the deduction. These knowledges and their interactions can also be represented by argumentation.

Initially, argumentation was the study within the field of philosophy and law. After argumentation was identified as a way to study defeasible reasoning [2], which is a field of KR, it led to many studies of argumentation in terms of AI. Eventually, this led to the emergence of many argumentation frameworks. One of the earliest argumentation framework defined is the Abstract Argumentation (AA) [3], and later on, another popular argumentation framework emerged, called Assumption-Based Argumentation (ABA) [4].

AA and ABA are quite related. AA is defined using a set of arguments and attack relationships among the arguments in the set. Meanwhile, in ABA, the arguments and attacks are derived from the (deductive system) rules, assumptions, and contraries; in which ABA building blocks (the rules, assumptions, and contraries) are “at a lower level of abstraction than those of AA” [5].

ABA has been studied to be applicable in supporting many fields. Fan, et al. [6] has studied the application of ABA for decision-making [7, 8] in medical literature field, in which they select the most relevant research paper for each patient, given the properties of a patient. Meanwhile, ABA has also been studied to be applied in multi-agent systems, supporting the dialogues among agents [9-11] and decision making of the agents in the system [7, 12]. In other fields, Matt, et al. [13] has studied ABA application in e-procurement, to compare and select the best suppliers of goods and services during procurement negotiation. Besides that, there is also a study by Fan and Toni [14] that studied ABA in the context of game theory, mapping games into argumentations and dialogues. Toni [15] has studied the application of ABA in ontological reasoning for the semantic web, and evaluating the exchange of information and opinions at for web-based social network. Fan and Toni [16], [17-19 have explored applying argumentation based explanations. ABA has also been used for activity recognition [20, 21].

The main goal of having AI to argue is to determine which argument is the best. Determining which argument is the "best" is quite subjective. Hence, several "semantics" have been defined over the years, for both AA and ABA. The semantics define how "good" a certain argument is compared to the other semantics. For instance, it is more desirable to have an argument that is not contradicting with itself (i.e. "conflict-free"). As stated by Toni [5], AA and ABA can be transformed into each other because they are an instance of each other. This implies that semantics defined for AA can be applied to ABA, and vice versa.

To determine whether an argument satisfies certain semantics may be computationally complicated [22-24]. Hence, many attempts have been made to create programs that efficiently compute the semantics given an argumentative framework; these programs are called *solvers*. ASPARTIX [25] is an AA solver using answer set programming (ASP) that supports a broad range of semantics. Meanwhile, jArgSemSAT [26] aims to implement AA efficiently, by re-implementing in Java ArgSemSAT which won 2$^{nd}$ place at the 2015 International Competition on Computational Model of Argumentation (ICCMA). TOAST [27], is another AA solver that uses ASPIC+ framework to represent the structure of arguments. Despite the abundance of AA solvers, there aren't many ABA solvers out there. Some of ABA solvers includes the tree-based solver *proxdd* [28]; and the graph-based solver *grapharg* [29], and *abagraph* [30]; all of them using Prolog language.

Having a web interface makes the project easier to access and users can try out the solver without investing time in setting up the solvers in local environments. There are a handful of AA solvers having a web interface to the solvers, such as ASPARTIX [25] and TOAST [27]. Meanwhile, there were no ABA solver with a web interface yet.

Implementing an ABA solver in a modern high-level programming language, such as Python, will make the program more accessible to more users. Furthermore, extending the tool with a web interface will indeed make the solver even more accessible to more users. Hence, there is an urgent need to implement an ABA solver in modern language and interface it to the web so more users are made aware of ABA.

## 1.1. Objective and Scope

The objective of this project is to implement an ABA solver in a modern programming language that performs reasonably well to regular users and create a web interface of the ABA solver so it can be accessible by the public. The programming language Python is chosen because of it is one of the most popular programming languages [31] and because of its abundance of community-supported packages available at Python Package Index (PyPI)[1].

The scope of this project includes the understanding of ABA and its semantics, analysis of the implementation of ABA solver & its web interface, techniques employed to optimize semantics computation, and evaluation of the program performance. Semantics covered in this project includes conflict-free, admissible, complete, grounded, ideal, and stable. The definition of each semantics will be covered in the subsequent sections. This project does not do an analysis of the complexity of the semantics computation because it has been extensively studied for both AA and ABA, see [22-24].

This report is organized as follows: Chapter 1 introduces the project, including project objective and scope. Chapter 2 discusses the background knowledge required for understanding the project, namely, defines the ABA argumentation framework, its components, and its semantics. Chapter 3 describes the methodology of the project, such as software requirements and project architecture. Chapter 4 discusses the implementation aspect of the project, in both back-end and front-end, including optimization techniques employed in the project. Experiments are designed and the results are discussed in Chapter 5. Finally, conclusions are drawn on Chapter 6.

---

[1] https://pypi.python.org/

# 2. Background

The field of argumentation, which analyses the represented knowledge, has been studied extensively since the introduction of *abstract argumentation* by Dung [3] in 1995. An abstract argumentation framework, as defined by Dung [3], represents the knowledge as a set of vertices, representing arguments, and the edges, representing the attacks among arguments. A more complex form of argumentation is called structured argumentation, where an argument can derive from the support of another argument, hence it is common to represent a single argument as a tree. [32]

In 1997, Bondarenko, et al. [4] introduced the concept of assumption-based argumentation (ABA) framework, which is a well-known type of structured argumentation [33]. Simply put, an assumption-based argumentation framework consists of:

- Set of symbols.
- Set of (inference) rules, in notation of $c \leftarrow s_1, \ldots, s_m$, for $m \geq 0$; where $s_1, \ldots, s_m$, is a set (possibly empty) of symbols supporting claim $c$. When $m = 0$ (i.e. $s_1, \ldots, s_m$ is empty), this rule represents a ground truth.
- Set of assumptions.
- Contraries of each assumption, which defines which arguments can attack this assumption.

An argument can be represented as a tree, constructed using the set of symbols and the set of inference rules. Meanwhile, the attack relationships can be defined as a set of assumptions and contraries of the assumption. Definition of argument and attack between argument is given in subsequent sections.

For this chapter, let's define the following scenario to be used as an illustration in the examples of ABA and its representations.

**Scenario 1**

Andy, a student in a certain university is having a dilemma when registering modules in the upcoming semester. There are two modules conflicting in lecture timing and hence he can only register either one of them. The first module, CZ5000 is a great module, because it is being taught by a great professor, and the module is also very popular among his circle of friends. The other module, CZ6000, is also another great module because the topic is popular in the industry right now, and Andy's final year project is related to the topics taught in this module. After reading online reviews about CZ6000, Andy found out that CZ6000 popularity in the industry is no doubt true.

Since arguments are essentially a claim, supported by another claim, Scenario 1 can be formalized into the following rules:

$cz5000_is_great \leftarrow cz5000_great_prof, cz5000_popular_on_friends;$
$cz6000_is_great \leftarrow cz6000_popular_in_industry, cz6000_related_to_fyp;$
$cz6000_popular_in_industry \leftarrow;$

Notice that the 3rd rule (CZ6000 being popular in industry) is represented as a ground truth, meaning that the claim is no doubt hold in the scenario.

Formally, an ABA framework is a tuple $\langle \mathcal{L}, \mathcal{R}, \mathcal{A}, \overline{\ \ } \rangle$ where:

- $\langle \mathcal{L}, \mathcal{R} \rangle$ is a deductive system, where $\mathcal{L}$ is the *language* and $\mathcal{R}$ is the set of *inference rules* in the form of $s_0 \leftarrow s_1, \ldots, s_m$, for $m > 0$ and $s_0, s_1, \ldots, s_m \in \mathcal{L}$;
- $\mathcal{A}$, where $\mathcal{A} \subseteq \mathcal{L}$ is a non-empty set, named the *assumptions*;
- $\overline{\ \ }$ is a total mapping from $\mathcal{A}$ to $\mathcal{L}$, where $\overline{a}$ is defined as the contrary of $a$.

The implication of this formal definition is that for every assumption that is present in the framework, it needs to have a contrary. A contrary defines which other argument can attack this assumption. The formal definition of attacks in ABA is presented in Chapter 2.2.

Using the definition of an ABA framework, Scenario 1 above can be represented as the following ABA framework:

$\mathcal{L} = \{$

$cz5000_is_great,$

$cz5000_great_prof,$

$cz5000_popular_on_friends,$

$cz6000_is_great,$

$cz6000_popular_in_industry,$

$cz6000_related_to_fyp$

$\}$

$\mathcal{R} = \{$

$cz5000_is_great \leftarrow cz5000_great_prof, cz5000_popular_on_friends;$

$cz6000_is_great \leftarrow cz6000_popular_in_industry, cz6000_related_to_fyp;$

$cz6000_popular_in_industry \leftarrow;$

$\}$

The above ABA framework is not yet complete as the assumptions and contraries are not yet defined, as this concept will be introduced in Chapter 2.2.

## 2.1. Arguments in ABA

Formally [34], given a deductive system $\langle \mathcal{L}, \mathcal{R} \rangle$, with language $\mathcal{L}$ and set of (inference) rules $\mathcal{R}$, and set of assumptions $\mathcal{A} \subseteq \mathcal{L}$, an *argument* for *claim* $c \in \mathcal{L}$ supported by $S \subseteq \mathcal{A}$ is a tree with nodes labelled by sentences in $\mathcal{L}$ or by symbol $\tau$, such that:

- The root is labelled by *c*
- For each node *N*
  - If *N* is a leaf, then *N* is labelled by either an assumption or by $\tau$
  - If *N* is not leaf, then there is an inference rule $l_N \leftarrow s_1, \ldots, s_m$ $(m \geq 0)$, where $l_N$ is the label of *N* and
    - If m = 0, then the rule shall be $l_N \leftarrow \tau$ (i.e. child of N is $\tau$)
    - Otherwise, *N* has *m* children, labelled by $s_1, \ldots, s_m$
- *S* is the set of all assumptions labelling the leaves

An argument [34] with claim *c* supported by a set of assumption *S* can also be denoted as $S \vdash c$.

Using Scenario 1, if the symbols $cz5000_great_prof$ and $cz5000_popular_on_friends$ are defined as assumptions, the following argument tree with claim $cz5000_is_great$ can be constructed.

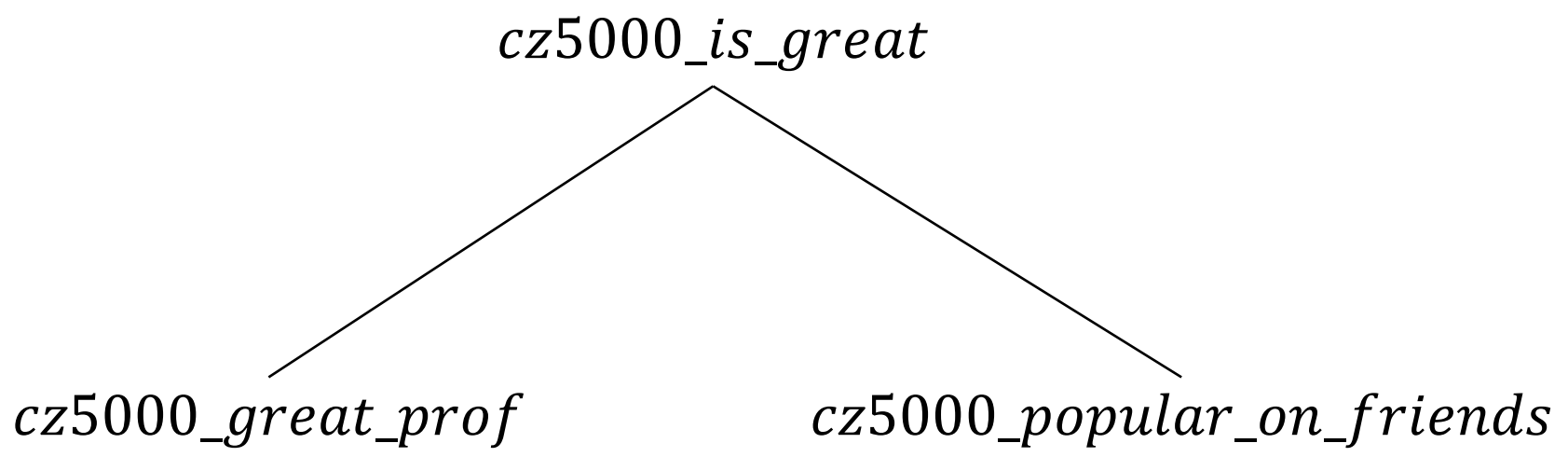

**Figure 1 Argument tree for *cz5000_is_great* on Scenario 1**

Also, if $cz6000_related_to_fyp$ is treated as assumption, the following argument tree with claim $cz6000_is_great$ can be constructed.

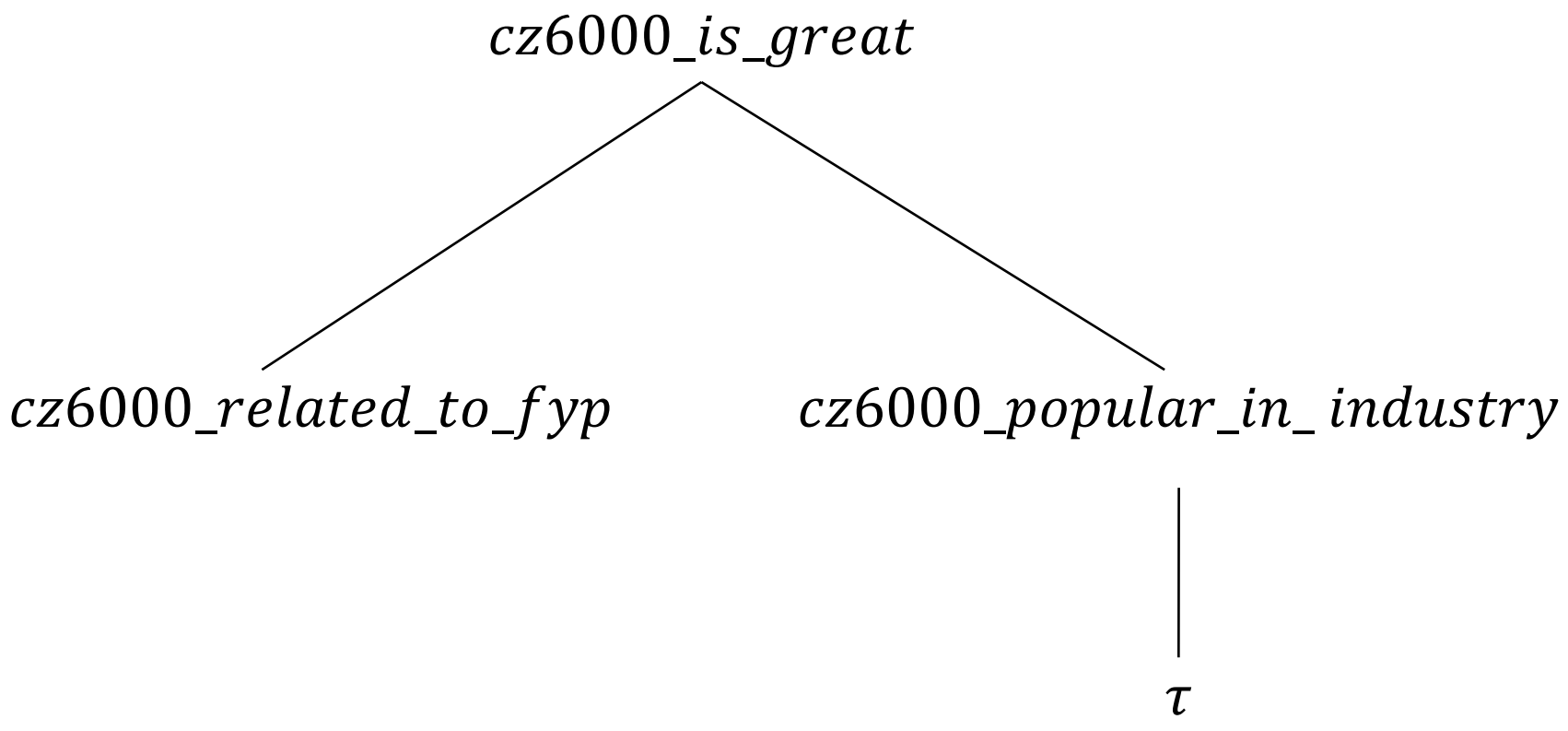

**Figure 2 Argument tree for *cz6000_is_great* on Scenario 1**

## 2.2. Attacks in ABA

Formally, an attack in ABA is defined as a relationship between two arguments $X$ and $Y$, where an argument $X$ *attacks* an argument $Y$ (written as $X \rightsquigarrow Y$) if and only if there are some $y \in S_Y$ (member of set of all assumptions supporting Y) and $\bar{y} = X$ (contrary of $y$ is the claim of argument $X$). In other words, an argument $A$ is said to be attacking another argument $B$ if there exists an assumption supporting $B$ which is the contrary of a sentence supporting $A$.

Going back to the recurring scenario, Andy is just informed that upon reading online reviews, he found out that the professor teaching CZ5000 is not that great as perceived by him. One review stated that the professor "conducts an extremely difficult final examination".

Taking the new information into account, one can construct the following contrary to be added to the pre-existing ABA framework.

$$\overline{cz5000_great_prof} \leftarrow cz5000_exam_is_hard$$

Note that this does not make the current ABA argument an actual argument yet (still potential argument), as not all the argument leaves are labelled as $\tau$ or assumption yet.

By that addition of contrary, there will be an attack relationship to the argument for $cz5000_is_great$ from the argument for $cz5000_exam_is_hard$.

## 2.3. Acceptability of Arguments

It is quite subjective to say whether an argument is the best argument, hence a less specific term "acceptability" is used. Whenever an argument is given, one can use ABA to determine whether it is "acceptable" or not. "Acceptability" of a claim is quite subjective, and hence can be formalized in many ways, defined as *semantics*.

There are six semantics that were implemented in this project. The formal definitions of the implemented semantics are given as follows:

1. Conflict-free: A set of arguments is conflict-free iff it does not attack itself. [30]
2. Admissible: A set of arguments is admissible iff it is conflict-free and attacks all arguments attacking it. [34]
3. Complete: A set of arguments is complete iff it is admissible and contains all arguments it can defend (by attacking every argument attacking them) [34]
4. Grounded: A set of arguments is grounded iff it is minimally (with respect to set inclusion) complete. [34]

5. Stable: A set of arguments is stable iff conflict-free and attacks every argument that does not belong to itself. [4]
6. Ideal: A set of arguments is ideal iff admissible and there are no admissible arguments attacking it. [35]

To compute the semantics, argument trees and *dispute trees* were required to be built. Informally, dispute tree can be understood as a mean to construct a winning strategy for a proponent's argument which is under dispute.[36] The proponent starts by presenting an initial argument. Then, proponents and opponents alternately attack each other's previous arguments. This continues infinitely or until there are no other arguments available to attack the proponents. In the latter case, it is said that the proponent wins the dispute. Hence, dispute tree displays the interaction of attacks among the arguments and derive whether a semantics hold based on the tree properties.

Formally, a *dispute tree*[34] for argument *A* is a tree $\mathcal{T}$, where:

1. Every vertex of tree $\mathcal{T}$ is either a proponent or opponent vertex (but not both), labelled by an argument.
2. Root vertex of tree $\mathcal{T}$ is a proponent vertex labelled by *A*.
3. For every proponent vertex *V* labelled by an argument *B* and for every argument *C* attacking *B*, there exists a child of *V*, which is an opponent labelled by *C*.
4. For every opponent vertex *V* labelled by an argument *B*, there exists exactly one child of *N* which is a proponent vertex *V* labelled by an argument attacking *B*.
5. There are no other vertices in tree $\mathcal{T}$ except those given by 1-4 above.

A branch of a dispute tree, i.e. a subtree, can have either finite or infinite depth. The infinite-depth tree can be caused by a circular attack. For example, in the following example, argument *A* attacks argument *B*, argument *B* attacks argument *C*, argument *C* attacks argument *A*, and later, argument *A* attacks argument *B*, and it goes on to infinity.

**Example 1**

$\mathcal{L} = \{a, b, c\}$
$\mathcal{R} = \{\,\}$
$\mathcal{A} = \{a, b, c\}$
$\bar{a} = b$
$\bar{b} = c$
$\bar{c} = a$

$A = \{a\} \vdash a$
$B = \{b\} \vdash b$
$C = \{c\} \vdash c$

In Example 1, arguments *A*, *B*, *C* are presented. The dispute tree of argument *A* is shown in the following figure.

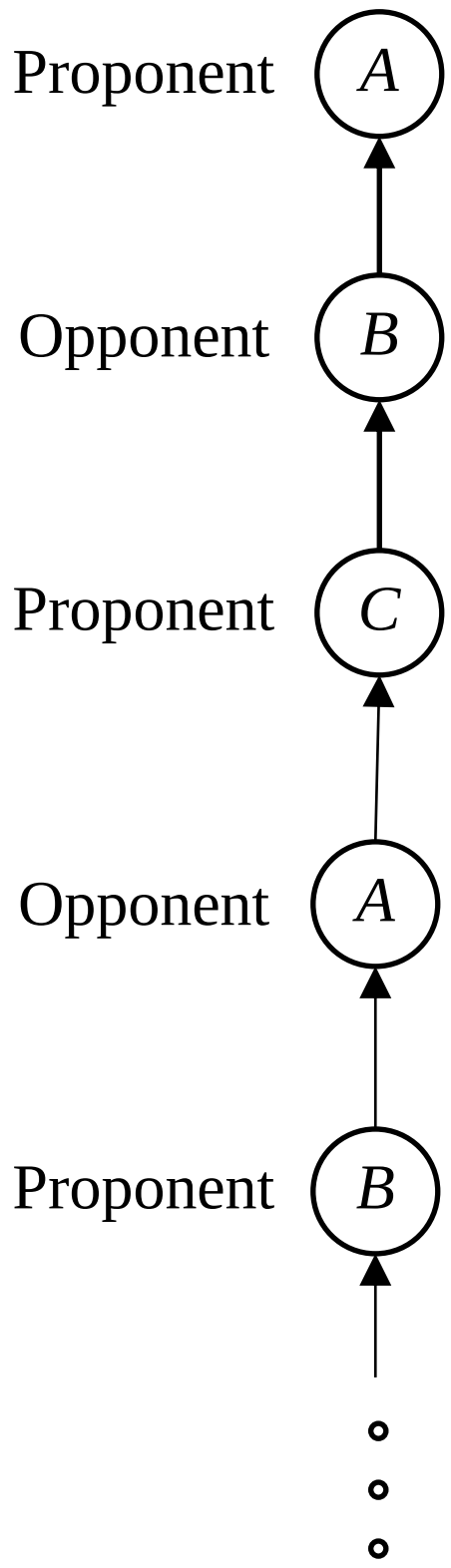

**Figure 3 Illustration of infinite dispute tree in Example 1**

From Figure 3, it is shown that the attacks are "circular" among argument *A*, *B*, and *C*; and will go on indefinitely. This situation illustrates an infinity dispute tree.

## 2.4. Computation of Acceptability

Given an ABA framework, its acceptability, defined in semantics, can be computed. In this section, the specifics of the semantics are described in more details.

### 2.4.1. Conflict-free

This is the simplest semantics to be computed. The real-life analogy is quite simple: it is undesirable to assert an argument that contradicts itself.

To check whether an argument is conflict-free, it is required that the argument tree has been constructed. After an argument tree is constructed, it is conflict free when the claim and assumption of all contraries do not belong to the tree.

---

**Example 2**

---

$\mathcal{L} = \{a, b\}$
$\mathcal{R} = \{a \leftarrow b;\}$
$\mathcal{A} = \{b\}$
$\bar{b} = a$

$A = \{b\} \vdash a$

---

In this example, argument *A* is defined as an argument with claim *a* supported by assumption *b*. Hence, one can construct an argument tree with *a* at its root, like in the following figure.

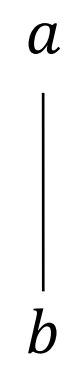

**Figure 4 Argument tree of Example 2**

But since argument *A* contains *a* that can attack *b* (because *b* is the contrary of *a*), hence it is said that argument *A* (with *a* as its claim) is not conflict-free.

### 2.4.2. Admissible

An argument is admissible if and only if the proponent and opponent in its dispute tree are consistent, meaning that, the proponent will always be proponent in the tree, and vice versa. A real-life example is that when arguing, a same kind of argument should be consistent, either attacking your argument or attacking your opponent, not switching sides in subsequent attacks.

The example presented at Example 1 is not an admissible because on its dispute tree, initially argument A is labelled as a Proponent, but subsequently, it is shown that argument A is being assigned "Opponent" status.

**Example 3**

$\mathcal{L} = \{a, b\}$
$\mathcal{R} = \{\,\}$
$\mathcal{A} = \{a, b\}$
$\bar{b} = a$
$\bar{a} = b$

$A = \{a\} \vdash a$
$B = \{b\} \vdash b$

In this example, check argument *A*'s admissibility, one need to build a dispute tree with argument *A* as its root argument.

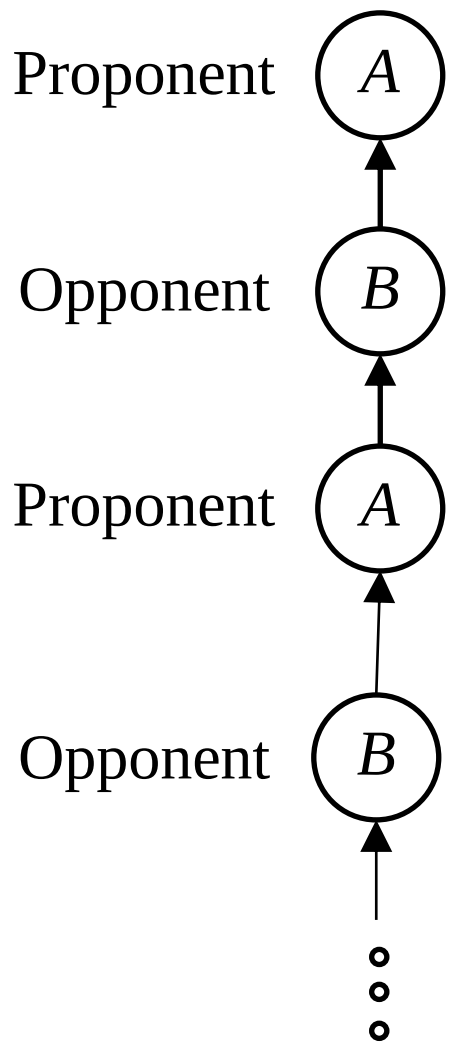

**Figure 5 Illustration of dispute tree of argument A in Example 3**

From the dispute tree shown in Figure 5, it is seen that while the tree is an infinite tree, the statuses of the arguments are consistent, argument *A* always having Proponent and argument *B* always having Opponent as their status. Hence, the argument *A* is admissible.

### 2.4.3. Complete

An argument is complete if and only if it is admissible, and it contains all arguments in which the attackers attacking them are attacked by this argument. An equivalent scenario is that one desires an argument that is a completely defendable argument, meaning that all assumptions supporting the parent argument, if being attacked, should be attacked by the parent argument.

Argument *A* of Example 3 is an example of a complete argument. This is because all its assumption ($a$) that can be attacked by others (argument *B*) is contained by itself (argument *A*).

### 2.4.4. Grounded

An argument is grounded if and only if the dispute tree is not infinite. Infinity of a tree branch is detected by tracking a history of what argument has been encountered so far. A grounded argument in real-life example is that when one can determine that the argument is the best whenever it has the final say, and not having endless attacks.

Argument *A* of Example 3 can be said as not grounded as its dispute tree, shown in Figure 5, is an infinite dispute tree.

---

**Example 4**

---

$\mathcal{L} = \{a, b\}$
$\mathcal{R} = \{b \leftarrow;\}$
$\mathcal{A} = \{a\}$
$\bar{a} = b$

$A = \{a\} \vdash a$
$B = \{\} \vdash b$

---

In this example, argument *B* is a ground truth. To illustrate, the argument trees of argument *A* and argument *B* are presented in the figure below.

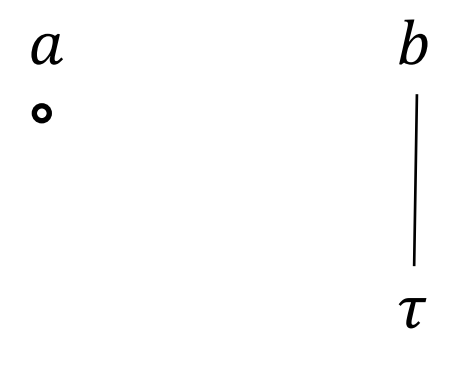

**Figure 6 Argument tree of Example 4**

To determine the grounded semantics, like admissible semantics, a dispute tree needs to be built. For example, to check argument $A$'s grounded semantics, one need to build a dispute tree with argument $A$ as its root argument.

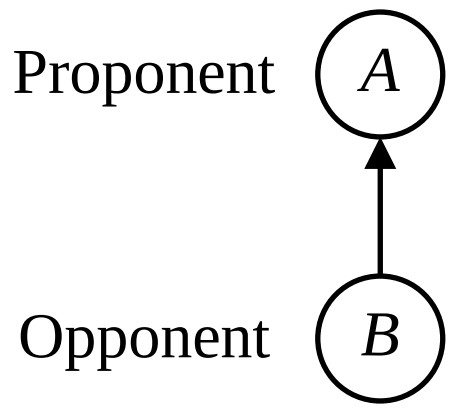

**Figure 7 Illustration of dispute tree of argument A in Example 4**

From Figure 7, it is shown that argument $A$ is grounded because the dispute tree is a finite dispute tree.

### 2.4.5. Stable

An argument is stable if and only if it is conflict-free and it attacks all assumptions not present. The implication of having a stable argument is that when one has a stable argument, it can attack everyone else's argument, and hence a stable argument looks "stronger" than anyone else's arguments.

**Example 5**

$\mathcal{L} = \{a, b, p, q\}$
$\mathcal{R} = \{a \leftarrow p; p \leftarrow; b \leftarrow q;\}$
$\mathcal{A} = \{q\}$
$\bar{q} = p$

$A = \{\} \vdash a$
$B = \{q\} \vdash b$

In this example, stable semantics is fulfilled by argument $A$ and it can be illustrated using argument trees below.

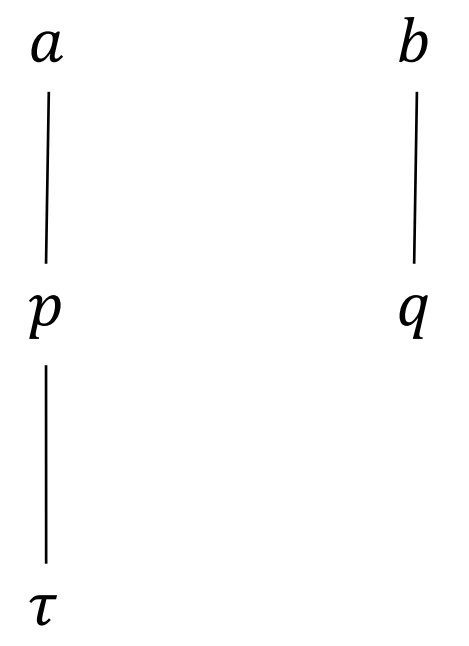

**Figure 8 Argument tree of Example 5**

From Figure 8, it is shown that argument *A* is conflict-free because there are no attacks from itself to members of its argument tree. Besides that, since argument *A* also attacks argument *B*, because *q* (supporting argument *B*) is the contrary of *p* (supporting argument *A*), argument *A* also fulfils stable semantics, since the argument attacks all assumptions (in the ABA) that do not present in its own argument.

### 2.4.6. Ideal

After the computation of admissible semantics has been carried out, the ideal semantics is computed. An ideal semantics is implemented quite straightforwardly: for each argument, get the attackers and check whether the attackers are admissible. If the attackers are admissible, then it is not an ideal semantics.

In life, ideal argument is when there are no admissible attackers; this made the argument looks stronger because the attackers are deemed "weaker" as they are not admissible.

# 3. Methodology

In order to follow a good software development practices, a software engineering process has been followed throughout the project as described by Bruegge and Dutoit [37]. In this report, to simplify the software engineering processes, only software requirements and software architecture were reported.

## 3.1. Software Requirements

### 3.1.1. Use Case Requirements

Use case requirements list down the overview scenario of when and how the program is being used. The use case of this project includes: computing ABA via web-interface and computing ABA via command line interface. For each use case, a detailed description is given. The use case description includes the use case name, usage frequency, priority, pre-conditions, post-conditions, flow of events, alternative flows, exceptions, etc.

### 3.1.2. Functional Requirements

Functional requirements of this project mainly cover the user computing an ABA framework via the web interface and computing ABA framework via the command line interface. Overall, the requirements cover the construction of argument trees, dispute trees, computing the semantics, and generating a visualization.

### 3.1.3. Non-functional Requirements

The non-functional requirements of this projects are mainly focused on the project performance, reliability, and maintenance. Performance requirement includes the need of the project being free from critical errors and being able to respond back within a reasonable amount of time. Reliability means that the program shall be available and ready to respond at any time. The program shall be designed such that it supports easy maintenance and easy bug-fixing in the future.

## 3.2. Software Architecture

In developing the project to satisfy all the software requirements, a good software architecture was designed. The project was fabricated in an object-oriented technique, where encapsulation and abstraction were applied. The other two principles, inheritance and polymorphism, were not implemented as there was no need to overcomplicate the project by implementing the whole object-oriented principles (per the KISS principle). DRY (don't repeat yourself) principle was also held in this project, where code duplication was aimed to be minimized as much as possible.

### 3.2.1. Class Diagram

In this project, the classes implemented and their interactions were captured in the class diagram below.

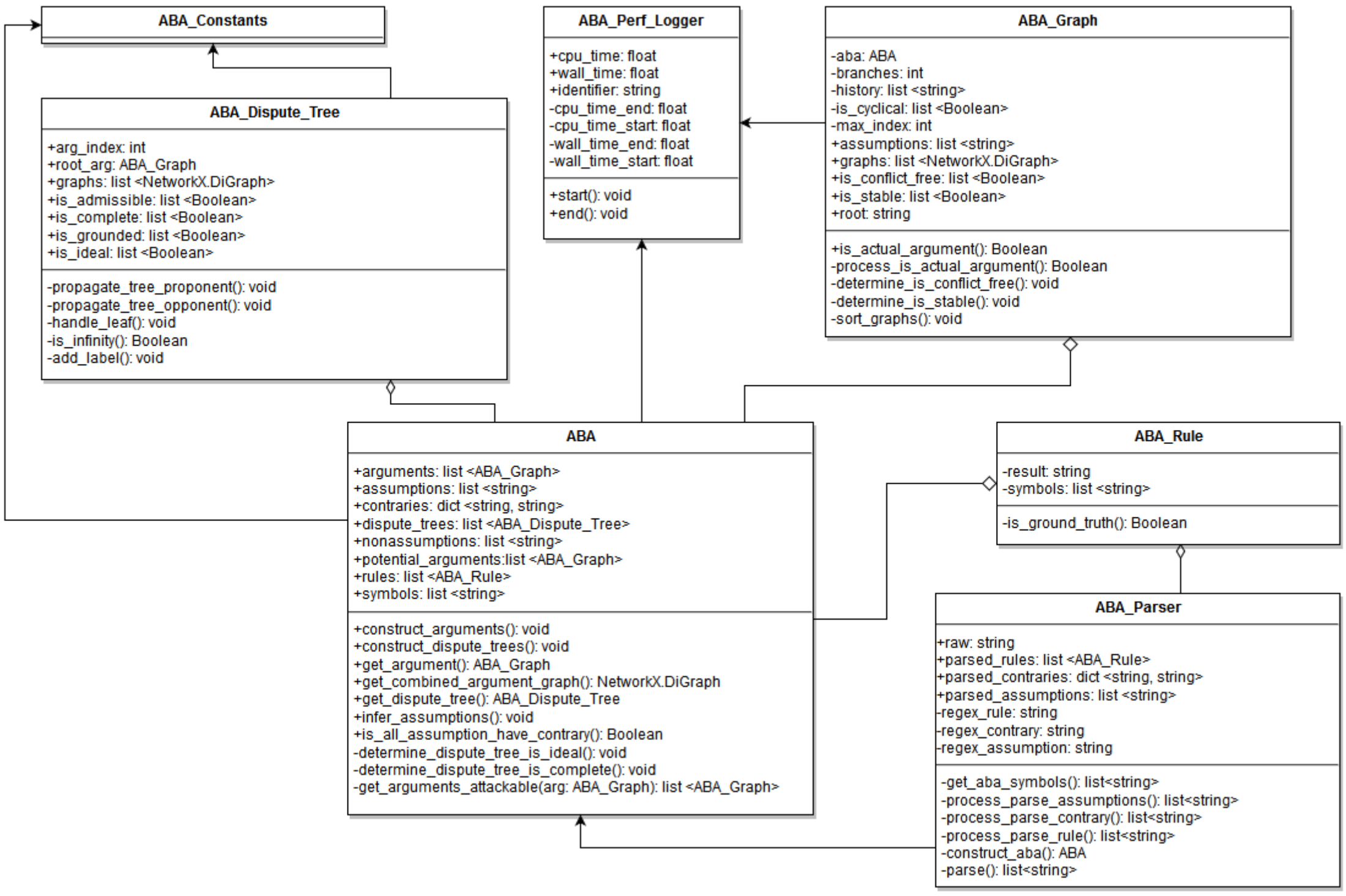

**Figure 9 Class Diagram**

### 3.2.2. Description of Classes

The description of each classes implementations is described in this section.

#### 3.2.2.1. ABA

This is the main class, containing the properties an Assumption-Based Argumentation framework, such as symbols, (inference) rules, assumptions, and contraries. If undeclared, assumptions can be derived from the contraries previously declared.

Constructing a minimal *ABA* class starts from making an instance of *ABA* class. The first thing to do is to declare the symbols that are present in the ABA at "*symbols*" property. After that, the "*rules*" property, which is a list, shall be filled with objects that are instance of *ABA_Rule* class. Then, the "*contraries*" property, which is a dictionary, shall be filled with the ABA's contraries containing the key-value pair: the key is a string representing the assumption symbol, and value is the symbol that attacks this assumption. After these minimal declarations, one can call "*infer_assumptions*" method, to extract the assumptions from the contraries, since every key at contraries is an assumption.

After that, one can call "*construct_arguments*" method to construct the argument trees. Internally, "*construct_arguments*" will check whether all assumptions have its contrary, and will append instances of *ABA_Graph* objects to "*potential_arguments*" property; and if the object satisfies the criteria being an actual argument, it will be appended to "*arguments*" property.

"*construct_dispute_tree*" method mainly generates instances of *ABA_Dispute_Tree* and append to "*dispute_trees*" property. After the generation process has been completed, it will determine whether the objects satisfy *ideal* and *complete* semantics.

#### 3.2.2.2. ABA_Rule

This class represents a single rule in ABA: a set of symbols (possibly empty) supporting a single symbol. This class has two public properties:

- "*symbols*", a list representing the symbols supporting the result of this rule
- "*result*", a string representing the result or claim or conclusion of this rule

If “*symbols*” is an empty list, the rule represents a ground truth. A helper public method “*is_ground_truth*” is available in this class that returns a Boolean value on whether the rule is a ground truth or not.

#### 3.2.2.3. ABA_Graph

This class represents arguments of the same claim symbol in ABA.

This class constructor requires an *ABA* object and a string representing argument claim symbol, in which the argument tree is to be constructed. The class constructor will call the construction of the argument trees, extraction of assumption set of each argument, and conflict-free and stable semantics computation for each argument.

This class contains the following public properties, which are available for usage after the class has been instantiated:

- “*root*”, a string, representing the root of the argument trees.
- “*graphs*”, a list containing *Networkx DiGraph* objects, each object representing an argument with the same root symbol.
- “*assumptions*”, a list of list containing the assumption set for each argument.
- “*is_conflict_free*”, a list representing whether the argument (of given index) satisfies conflict-free semantics.
- “*is_stable*”, a list representing whether the argument (of given index) satisfies stable semantics.

This class only contain one public method “*is_actual_argument*” that determines whether the argument (of a given index) is an actual argument.

#### 3.2.2.4. ABA_Dispute_Tree

This class represents dispute tree of each argument in ABA. This represents one or possibly more dispute tree generated from one *ABA_Graph* object. This dispute tree is needed to compute the grounded and admissible semantics.

The public properties of this class, in which all but two are available after class instantiation, are:

- “*graphs*”, a list of *Networkx’s DiGraph* objects, representing the dispute trees with the same argument root.
- “*root_arg*”, an *ABA_Graph* object, representing the argument at the root of dispute tree.
- “*arg_index*”, an integer, representing the argument tree’s index at the *ABA_Graph* object. Since an *ABA_Graph* object might have more than one argument, it is required to have a tracking integer to point which argument that is currently being used for constructing the dispute tree.
- “*is_grounded*”, a list representing whether the dispute tree (of given index) satisfies grounded semantics.
- “*is_admissible*”, a list representing whether the dispute tree (of given index) satisfies admissible semantics.
- “*is_complete*”, a list representing whether the dispute tree (of given index) satisfies ideal semantics. This property is externally computed at *ABA* class because it needs to know other dispute trees admissible semantics.
- “*is_ideal*”, a list representing whether the dispute tree (of given index) satisfies ideal semantics. This property is externally computed at *ABA* class because it needs to know other dispute trees admissible semantics.

There are no public methods available on this class. The private methods available in this class are used to recursively construct the dispute tree, simulating the attacks between the proponent nodes and opponent nodes.

#### 3.2.2.5. ABA_Parser

This class is a helper class to parse raw text and returns an *ABA* object. This class is meant to be the interface between the web interface and the rest of the project, as this class parses the rules from raw string and returns an *ABA* object, after handling symbols inference, assumptions inference, arguments construction, dispute tree generations, and semantics computations.

Public properties of this class are as follows:

- “*raw*”, a string, containing the unparsed text of *ABA_Parser* input.

- "*parsed_rules*", a list containing *ABA_Rule* objects, representing the list of rules parsed from the raw string. Property value is available after calling "*parse*" method.
- "*parsed_contraries*", a dictionary (key-value pair), where the key is a string representing assumption and value is a string representing symbol that attacks this assumption. This property is parsed from the "*raw*" string, hence the value is available after calling "*parse*" method.
- "*parsed_assumptions*", a list of string representing the assumptions explicitly declared at the "raw" string. Assumption inference from contraries is not done here. Property value is available after calling "*parse*" method.

Public methods of this class are as follows:

- "*parse*". At a high level, this method will check for parsing errors and try to parse the "raw" property to "*parsed_rules*", "*parsed_contraries*", and "*parsed_assumptions*". List of errors encountered during parsing is returned from this method call.
- "*construct_aba*". This is a helper method to return an *ABA* object. Internally, this method will initialize an *ABA* object, and then construct its class properties, such as "*symbols*", "*rules*", "*assumptions*", and "*contraries*". It will also do symbols inference from rules, and assumption inference from contraries. After that, it, too, will call for the construction of ABA argument trees and ABA dispute trees. Finally, it will return a fully constructed *ABA* object, where the argument trees, dispute trees, and semantics are fully computed and directly usable.

#### 3.2.2.6. ABA_Constants

This class is a helper class to store some constants used in the computations. Constants that are stored here currently includes the string for *DT_PROPONENT* (the status string for proponent dispute tree node), *DT_OPPONENT* (the status string for opponent dispute tree node), and *DT_MAX_BRANCH* (the maximum branching of a dispute tree).

#### 3.2.2.7. ABA_Perf_Logger

This class is a helper class to perform performance logging. Logging is stored on the same file specified at the program runtime.

Public properties available on this class are:

- "*wall_time*", this contains floating point number representing the time taken, in seconds, for the process to run, as perceived by human user. The measured time is the time between calling "*start*" and "*end*" methods. The number is only available after calling "*end*" method.
- "*cpu_time*", this contains floating point number representing the time taken, in seconds, for the process to be utilized in the CPU. The measured time is the time between calling "*start*" and "*end*" methods. The number is only available after calling "*end*" method.
- "*identifier*", a string to identify the current logging case

Public methods available on this class are:

- "*start*". Internally, it will record down the current time for later usage at "*end*" method.
- "*end*". Internally, it will record down the current time and compare it to the time recorded during "*start*" method. After that, the difference between those values is stored at "*wall_time*" and "*cpu_time*" class properties. Finally, this method will write the time measured values into the log file.

# 4. Implementation

This project is divided into two main areas: back-end computation and front-end web interface.

## 4.1. Back-end

Back-end computation was implemented in Python (version 3.4) using no framework but in object-oriented programming. Test-driven development was observed from the beginning of the project, where test cases were implemented first before the actual implementation. Thirty-three test cases were implemented in Python's native unit test framework. Source code tracking and management was done using Git and is published at GitHub[2].

Computation of the semantics are mainly using the dispute tree approach as described in Dung, et al. [34] and Dung, et al. [35]. The argument trees and dispute trees are implemented using the help *Networkx*[3] (Python) library.

The following lists the overview technical implementation of each semantics. Performance evaluations will be presented in the subsequent chapter.

- Conflict-free: After an argument tree is constructed, it is conflict free when both the attacker and assumption of all contraries present in the ABA do not belong to the tree.
- Admissible: An argument is admissible if and only if the proponent and opponent status in its dispute tree are consistent throughout the tree construction, meaning that, the proponent will always be proponent in the tree, and vice versa.
- Complete: An argument is complete if and only if it is admissible, and it contains all arguments in which the attackers attacking them are attacked by this argument.

[2] Available at https://github.com/kenrick95/aba-web.
[3] Available at PyPI: https://pypi.python.org/pypi/networkx

- Grounded: An argument is grounded if and only if the dispute tree is not infinite. Infinity of a tree branch is detected by tracking a history of what argument has been encountered so far.
- Stable: An argument is stable if and only if it is conflict-free and it attacks all assumptions not present in itself.
- Ideal: An ideal semantics is implemented quite straightforwardly: for each argument, get the attackers and check whether the attackers are admissible. If the attackers are admissible, then return false.

This section discusses the implementation of construction of argument trees, dispute trees, and each computation of the semantics.

Note that the procedures shown here are simplified versions of what was implemented in the project and some functionalities have been stripped away so the main gist of the implementation can be understood. In the procedures shown here, it is assumed that there is at most one rule with the same claim in the ABA, meaning that there cannot be two rule supporting the same symbol (e.g. $a \leftarrow p;$ and $a \leftarrow q;$ does not exist) in the ABA. In the true implementation of this project, to have more than one rule supporting a same claim, recursive branching is implemented for each rule. For example, in the case of $a \leftarrow p;$ and $a \leftarrow q;$, there will be two argument trees with $p$ as its claim since it can be constructed with either $a \leftarrow p;$ or $a \leftarrow q;$.

#### 4.1.1. Argument tree

The following procedure describes the construction of an argument tree, given an ABA framework and a symbol labelling the claim of the argument.

---

**Procedure 1.1.**: Constructing argument tree

---

<u>function constructArgumentTree </u>(*rootSymbol*, *aba*):
**Input**:
    *rootSymbol*: a string, representing the symbol labelling the root/claim/conclusion of the argument tree.
    *aba*: an ABA object, describing the properties of the ABA framework.
**Output**:
    An argument tree.

*argumentTree* = object();
*argumentTree*.rootSymbol = *rootSymbol*;

```
argumentTree.graph = NetworkX.DiGraph();
argumentTree.is_cyclic = false;
history = list ();

argumentTree.graph.addNode(rootSymbol);

history.append (rootSymbol);
expandNode (rootSymbol, argumentTree, aba, history);
history.pop ();

return argumentTree;
```

Procedure 1.1. describes the initialization of an argument tree, but the construction would not be complete without Procedure 1.2. below, which recursively expands the nodes to find the node supporters.

**Procedure 1.2.**: Argument tree expansion

```
function expandNode (node, argumentTree, aba, history):
Input:
        node: a string, representing the symbol that will be expanded.
        argumentTree: an argument tree
        aba: an ABA object, describing the properties of the ABA framework.
        history: a list of previously seen symbols before expanding the current node.

for rule in aba.rules do
        if rule.result == node then
                for symbol in rule do
                        argumentTree.graph.addEdge (node, symbol);
                        if symbol is not None then
                                if symbol in history then
                                        argumentTree.is_cyclic = true;
                                        break;
                                end if
                                history.append (symbol);
                                expandNode (symbol, argumentTree, aba, history);
                                history.pop ();
                        end if
                end for
        end if
end for

return argumentTree;
```

Expanding an argument tree is straightforward. Firstly, from the rules available in the ABA framework, only selects those that derives the current node symbol, i.e. the current node symbol is the result/claim/conclusion of the rules. Using these rules, an edge is added between the node and the supporting symbols present in the rule. If the supporting symbols are not *None*, meaning that it is not ground truth, then expand it further. Before expansion, it is crucial to check whether the tree built is cyclic, because the argument tree that is being built from this process does not guarantee that it is an actual argument, hence there will be another function to check whether an argument tree is an actual argument or just a potential one. Checking whether an argument tree is cyclic can be done by maintaining a list of symbols seen so far in the node expansion, i.e. a "history". If a symbol that is about to be expanded has been seen, the expansion will be stopped, to avoid infinite recursion.

Note that this way of expansion will treat all rules of the same symbol as if they are one and all the supporting symbols in each rule are combined into one. This will make the computation wrong, as there is no concept of "or" rules in here. The way to handle this "or" rules is to "branch" out whenever there might be multiple rules having the same claim symbol. They way to "branch" out makes the subsequent procedures complicated and hence it is implemented in the project, but not shown in the report.

After argument tree construction has been completed, it is also important to check whether the tree is an actual argument or not, meaning that it abides with the definition of an argument tree. The checking is described in Procedure 1.3. and Procedure 1.4. below.

---

**Procedure 1.3.**: Determining whether an argument tree is an actual argument

---

function isActualArgument (*argumentTree, aba*):
**Input**:
    *argumentTree*: an argument tree
    *aba*: an ABA object, describing the properties of the ABA framework.
**Output**:
    Boolean value, whether the given *argumentTree* is an actual argument.

*isActual* = false*;*
**if** *argumentTree*.isCyclical **then**
    return false;
**end if**
return isBranchActual (*argumentTree*.rootSymbol, *argumentTree*, aba);

---

**Procedure 1.4.**: Determining whether an argument subtree is an actual argument

function isBranchActual (*node*, *argumentTree*, *aba*):
**Input**:
    *node*: the node in the argument tree that marks the root of the subtree
    *argumentTree*: an argument tree
    *aba*: an ABA object, describing the properties of the ABA framework.
**Output**:
    Boolean value, representing whether the given *argumentTree* is an actual argument.

*successors* = *argumentTree*.getSuccesors(*node*);
**if** count (*successors*) == 0 **then**
    **if** (*node* is $\tau$) or (*node* in *aba*.assumptions) **then**
        return true;
    **end if**
    return false;
**end if**
**for** successor in successors **do**
    *successorActual* = isBranchActual (*successor*, *argumentTree*, *aba*);
    **if** *successorActual* is false **then**
        return false;
    **end if**
**end for**

return true;

Determining whether an argument tree is an actual tree is a recursive process. First, if one argument "tree" has a cycle inside it, by definition of a tree, it is not an actual argument anymore. If an argument tree is an actual argument, the subtrees are also actual arguments. By using this lemma, recursive search for a non-actual argument can be employed. This procedure checks where all the leaves are labelled as either "$\tau$" or assumption. If there are violation of this rule, the whole tree will not be an actual argument.

#### 4.1.2. Dispute tree

Procedure 2.1. below constructs a dispute tree and on the process, it also computes the *complete* and *grounded* semantics. Dispute tree construction is a recursive process, where proponent nodes will be attacked by opponent nodes (Procedure 2.2.) and vice versa (Procedure 2.3.). In the tree construction, it will need to check whether the branch

has gone into an infinite branch or not (Procedure 2.4.), hence determining whether the tree is grounded or not; and whether the node status has changed during the recursion or not (Procedure 2.5.), hence determining whether the tree is admissible or not.

---

**Procedure 2.1.**: Construct dispute tree and determine *admissible* and *grounded* semantics

---

function constructDisputeTree (*rootArgument*, *aba*):
**Input**:
*rootArgument*: an argument tree, representing the initial argument in the dispute tree.
*aba*: an ABA object, describing the properties of the ABA framework.
**Output**:
whether the given argument, represented by the dispute tree satisfies complete semantics in the given ABA framework.

*disputeTree* = object();

*disputeTree*.graph = NetworkX.DiGraph();
*disputeTree*.isGrounded = true;
*disputeTree*.isAdmissible = true;
*history* = list ();

*disputeTree*.graph.addNode(*rootArgument*);
addStatus (*rootArgument*, *disputeTree*, “Proponent”);

*history*.append (<*rootArgument*, “Proponent”>);
expandProponentNode (*rootArgument*, *disputeTree*, *aba*, *history*);
*history*.pop ();

return *disputeTree*;

---

Dispute tree is implemented in this project as an object consisting of a *Networkx*’s graph object (representing the tree structure; it is a technical limitation that a directed graph is used instead of tree), and two other Boolean value properties, representing whether the dispute tree is grounded and admissible. Initially, both grounded and admissible semantics are assumed to hold until found otherwise. History is a variable to keep track of previously seen arguments during the tree expansion.

**Procedure 2.2.**: Expand dispute tree's proponent node

function expandProponentNode (*node*, *disputeTree, aba*, *history*):
**Input**:
    *node*: an argument tree
    *disputeTree*: a dispute tree
    *aba*: an ABA object, describing the properties of the ABA framework.
    *history*: a list of previously seen arguments before expanding the current node

**for** *assumption*, *symbol* in *node*.assumptions.items() **do**
    *opponentNode* = *aba*.getArgument(*symbol*);
    **if** *opponentNode* is None **then**
        **continue**

    *disputeTree*.graph.addEdge (*node*, *opponentNode*);
    addStatus (*opponentNode*, "Opponent", *disputeTree*);

    **if** isInfinity (*opponentNode, history*) **then**
        **break**;
    **end if**

    *history*.append (<*opponentNode*, "Opponent">);
    expandOpponentNode (*opponentNode, disputeTree, aba, history*);
    *history*.pop ();
**end for**

addToCache(*node, disputeTree*);

return**;**

Expansion of a proponent node is simply getting the assumption set of that argument, and for each assumption, get the argument that attacks it. For each argument that attacks this node, i.e. the *opponentNode*s, add an edge between the proponent and opponent nodes, symbolizing an attack, and then add the status of the *opponentNode* as "Opponent" in this dispute tree. Then do a check if the tree is infinity, if not, then expand the *opponentNode*. Finally, as an optimization technique, once the expansion of the *opponentNode*s has been completed, i.e. the branch has been fully grown, add the branch to the cache.

**Procedure 2.3.**: Expand dispute tree's opponent node

```
function expandProponentNode (node, disputeTree, aba):
Input:
        node: an argument tree
        disputeTree: a dispute tree
        aba: an ABA object, describing the properties of the ABA framework.

for assumption, symbol in node.assumptions.items() do
        proponentNode = aba.getArgument(symbol);
        if proponentNode is None then
                continue

        if existsInCache(proponentNode) then
                cachedBranch = loadBranchFromCache(proponentNode);
                NetworkX.mergeGraph(disputeTree.graph, cachedBranch);
        end if

        disputeTree.graph.addEdge(node, proponentNode);
        addStatus (proponentNode, "Proponent", disputeTree);

        if isInfinity (proponentNode, history) then
                break;
        end if

        history.append (<proponentNode, "Proponent">);
        expandProponentNode (proponentNode, disputeTree, aba, history);
        history.pop ();

        break;
end for

return;
```

The procedure of expansion of opponent node is quite similar to that of the proponent node. The first difference is that there is exactly one proponent node that attacks this node. Another difference is the optimization technique when it is about to expand the proponent node, which attacks this opponent node, it checks the cache if there is already a branch with that proponent node has already been fully grown. If so, then the branch us copied out from the cache and merged to the current dispute tree.

**Procedure 2.4.**: Determining whether a branch is infinite

function isInfinity (*node*, *history*):
**Input**:
*node*: an argument tree
*disputeTree*: a dispute tree
*aba*: an ABA object, describing the properties of the ABA framework.

*isInfinity* = <*node*, *status*> present in *history*;
**if** *isInfinity* **then**
*disputeTree*.isGrounded = false;
**end if**

return *isInfinity*;

Determining whether a branch is infinite or not is easily done by checking whether the given tuple <*node*, *status*> has already been present in the history. If so, then it is guaranteed that the tree expansion will cause a cycle, hence an infinite branch. This is because if a given tuple <*node*, *status*> has already been present in history, the route of node expansion will be the same as those present in *history* since the attackers of the argument *node* (and the subsequent descendants) will be the same as before. By definition, when a branch is seen to be infinite, the whole tree is said to be not grounded.

**Procedure 2.5.**: Add a status to a node, either "Proponent" or "Opponent"

function addStatus (*node*, *status*, *disputeTree*):
**Input**:
*node*: an argument tree
*status*: either "Proponent" or "Opponent"
*disputeTree*: a dispute tree

**if** *disputeTree*.nodes[*node*].status != *status* **then**
*disputeTree*.isAdmissible = false;
**end if**
*disputeTree*.nodes[*node*].status = *status*;

return;

Whenever there are new arguments being added to the dispute tree, a status is being assigned to them, they are either a proponent node or opponent node, but never both. Hence, this can be implemented by a dictionary (key-value pair) data structure, where the key is the argument (can be represented by a node) and the value is the status of the node (a string "Proponent" or "Opponent"). When the key has already existed and the value being written attempts to change the value to a different one, the dispute tree is being marked as not admissible.

Since a branch of a dispute tree can be either finite or infinite, it is essential to check whether a branch has gone into infinity, i.e. a "cycle" of attacks. Since a tree structure does not permit a cycle, the project employed a workaround by implementing the tree not as using tree data structure but using directed graph data structure. A cycle of attacks can be easily detected during the construction of the dispute tree by maintaining a stack of previously seen arguments (named "history"), that is an argument is pushed into the stack when expanding the node and popped when the expansion has finished. This way, whenever a new argument is being pushed into the history and is previously seen, it is said that infinity has occurred. Since the program main goal of dispute tree in the context of this project is to determine whether an argument is admissible or grounded, it can terminate from computation, because an infinity branch makes the argument in dispute tree, by definition, not grounded.

#### 4.1.3. Conflict-free

Determining whether an argument satisfies a conflict free argument is very simple, with the help from *Networkx* graph library. The following shows the pseudocode for determining whether a graph is conflict-free

---

**Procedure 3**: Determining whether an argument satisfies conflict-free semantics in a given ABA framework

---

function isConflictFree (*argumentTree*, *aba*):
**Input**:
    *argumentTree*: an argument tree.
    *aba*: an ABA object, describing the properties of the ABA framework.
**Output**:
    whether the given argument tree is conflict-free in the given ABA framework.

*conflictFree* = true
**for** (*assumption*, *attacker*) in *aba.contraries*.items()
    **if** *attacker* in nodes(*argumentTree*)
    and *assumption* in nodes(*argumentTree*) **then**
        *conflictFree* = false
        **break**
    **end if**

**end for**
return *conflictFree*;

An argument, represented by the argument tree, is being assumed as *conflict-free* until being found otherwise. The implementation of this project is simply getting the list of contraries of the ABA framework, and for each contrary (assumption-attacker pair), if both attacker and assumption belong to this argument tree, the argument tree is marked as not conflict free.

#### 4.1.4. Admissible

Admissibility of an argument is determined from its dispute tree during its construction phase, as described in Procedure 2.1. to Procedure 2.5. Initially, a dispute tree is assumed to be *admissible* until proven otherwise, i.e. it is henceforth determined as not admissible if and only if an argument is seen to have both proponent and opponent labelling that argument. (Procedure 2.4.)

#### 4.1.5. Complete

The pseudocode, given below, to determine whether an argument satisfies *complete* semantics or not is derived from its formal definition: An argument is complete if and only if it is admissible, and it contains all arguments in which the attackers attacking them are attacked by this argument.

**Procedure 4**: Determining whether an argument satisfies complete semantics in a given ABA framework

function isComplete (*disputeTree*, *aba*):
**Input**:
*disputeTree*: a dispute tree.
*aba*: an ABA object, describing the properties of the ABA framework.
**Output**:
whether the given argument, represented by the dispute tree satisfies complete semantics in the given ABA framework.

*complete* = false
**if** isAdmissible (*disputeTree*, *aba*) **then**
*complete* = true
**if** not isGrounded (*disputeTree*, *aba*) **then**
*attackablesByRoot* = getArgumentsAttackable (*disputeTree*. rootArgument, *aba*)

*defendableArguments* = []

```
        for attackable in attackablesByRoot do
                defendables = getArgumentsAttackable (attackable, aba))
                defendableArguments.extend(defendables)
        end for

        allInArgument = true
        for argument in defendableArguments do
                if argument.claim not in
nodes(disputeTree.rootArgument.tree) then
                        allInArgument = false
                        break
        end for
        complete = allInArgument;
return complete;
```

Since determining this depends on the knowledge of *admissible* semantics, hence it can only be computed only when the dispute tree has been fully constructed and admissibility semantics have been determined. Meanwhile, Craven and Toni [30] mentioned that if an argument has satisfied *grounded* semantics, it is guaranteed that it also satisfies *complete* semantics. Because at the dispute tree, both *admissible* and *grounded* semantics are determined, one can use these two pieces of knowledge to optimize the computation.

Proceeding to compute the complete semantics, first, the procedure gets a list of all arguments that can be attacked by the root (*attackablesByRoot*). Then, it constructs another list, which all members of *attackablesByRoot* can attack (*defendableArguments*). After that, it is said the root argument of this dispute tree is *complete* if and only if all members of *defendableArguments* are inside the argument tree of dispute tree's root argument.

#### 4.1.6. Grounded

The *grounded* semantics of an argument is determined from its dispute tree during its construction phase, as described in Procedure 2.1. to Procedure 2.5. Initially, like admissible semantics, a dispute tree is assumed to be *grounded* until proven otherwise, i.e. it is henceforth determined as not *grounded* if and only if the *grounded* tree become an infinite dispute tree, that is a "cycle" is found in the tree. (Procedure 2.5.)

### 4.1.7. Stable

The following procedure determines whether an argument satisfies *stable* semantics, given an ABA framework and its argument tree.

---

**Procedure 5**: Determining whether an argument satisfies stable semantics in a given ABA framework

---

function isStable (*argumentTree, aba*):
**Input**:
    *argumentTree*: an argument tree.
    *aba*: an ABA object, describing the properties of the ABA framework.
**Output**:
    whether the given argument satisfies stable semantics in the given ABA framework.

*stable* = false
**if** isConflictFree (*argumentTree, aba*) **then**
    *stable* = true
    **for** (*assumption, attacker*) in *aba*.contraries.items() **do**
        **if** *assumption* not in nodes(*argumentTree*) **then**
            **if** *attacker* not in nodes(*argumentTree*) **then**
                *stable* = false
                **break**
            **end if**
        **end if**
    **end for**
return *stable*;

---

An argument satisfies *stable* semantics iff it satisfies these two conditions:

1. The given argument is *conflict-free*. This is computed from the function *isConflictFree* from section 4.1.3
2. The given argument attacks all other arguments not supporting itself. The logic is simply iterating over the contraries (assumption-attacker pair), where if there are assumptions that are not a member of nodes in the argument tree, then the attacker of those assumptions must be a member of nodes in the argument tree, otherwise, the argument is said to be not *stable*.

### 4.1.8. Ideal

The following procedure determines whether an argument satisfies *ideal* semantics, given an ABA framework and its dispute tree.

**Procedure 5**: Determining whether an argument satisfies stable semantics in a given ABA framework

```
function isIdeal (disputeTree, aba):
Input:
        disputeTree: a dispute tree.
        aba: an ABA object, describing the properties of the ABA framework.
Output:
        whether the given argument, represented by the given dispute tree, satisfies
ideal semantics in the given ABA framework.

ideal = false
if isAdmissible (disputeTree.rootArgument, aba) then
        ideal = true
        for node in nodes(disputeTree) do
                if node.status is "Opponent" then
                        opponentDisputeTree = getDisputeTree (node, aba);
                        if isAdmissible (opponentDisputeTree, aba) then
                                ideal = false;
                                break;
                        end if
                end if
        end for
return ideal;
```

For an argument to satisfy the *ideal* semantics, by definition in [35], the dispute tree needs to satisfy the following requirements:

1. The dispute tree is *admissible*. This information is computed during the construction of dispute tree, see section 4.1.2 and 4.1.4, and hence the values can be used in this case.
2. There is no opponent in the dispute tree that satisfies *admissible* semantics. This is done by simply iterating over the nodes present in the dispute tree and checking the opponent nodes, whether these nodes are admissible or not. If there exists an opponent node that is *admissible*, it is said that the current dispute tree is not *ideal*.

## 4.2.Front-end

Front-end computation was implemented in Hypertext Mark-up Language (HTML), JavaScript (JS), and Cascading Style Sheet (CSS) with Flask (Python) micro-service as the connector from the web interface to the back-end. The web interface contains:

- Input area: the user should input using a simple Prolog-like syntax to define the ABA inference rules. (Figure 10)
- Visualization output: output is rendered using HTML's Scalable Vector Graphics (SVG) using D3.js library for both argument trees (Figure 10) and dispute trees (Figure 11). The back-end returns the argument trees and dispute trees in JSON format to the front-end using the help of *jsonpickle* (Python) library.
- Help page: listing the project features, input syntax, and limitations. This help page is served to guide users in using the project.

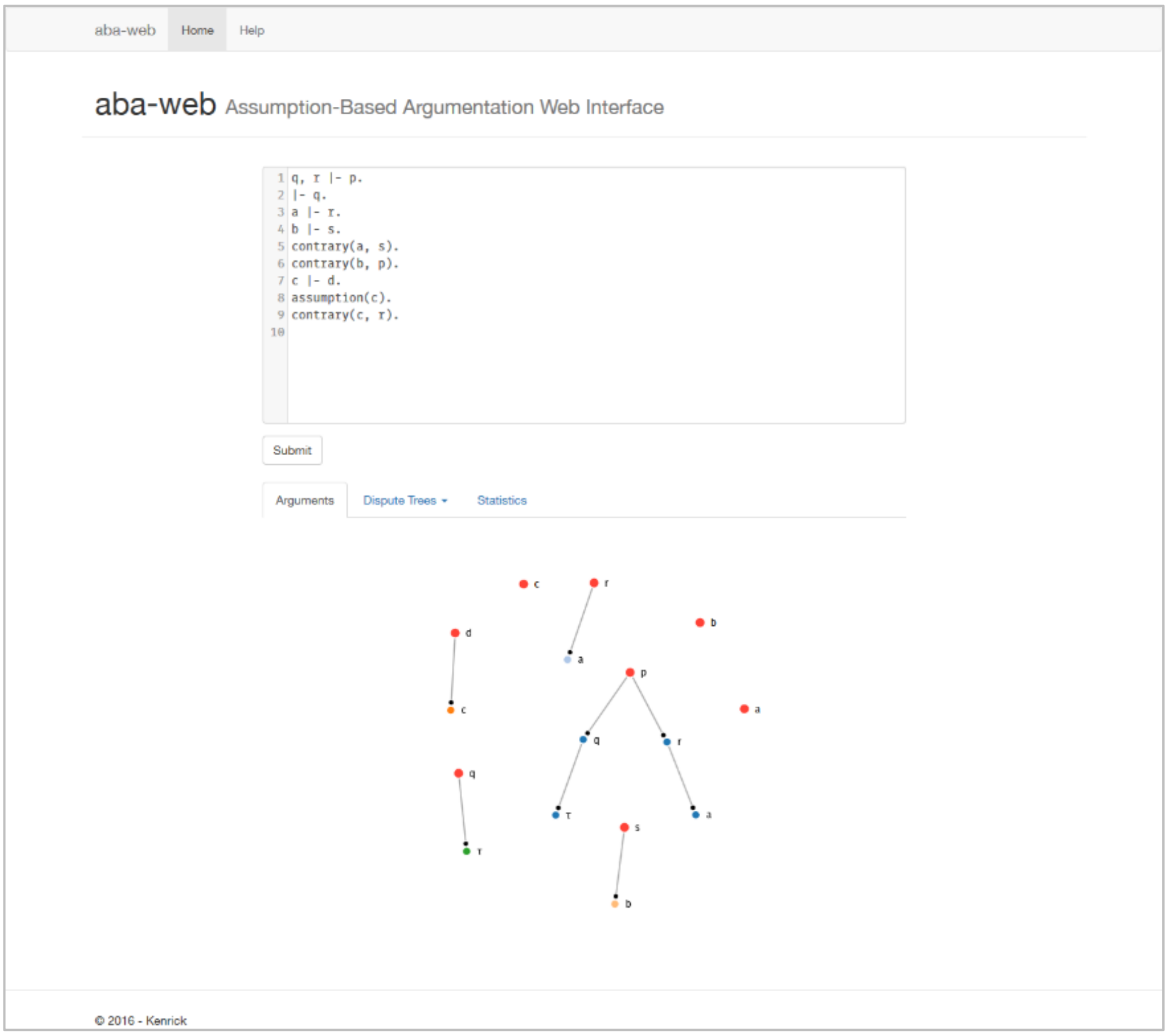

**Figure 10 Project web interface, showing the input area and argument trees**

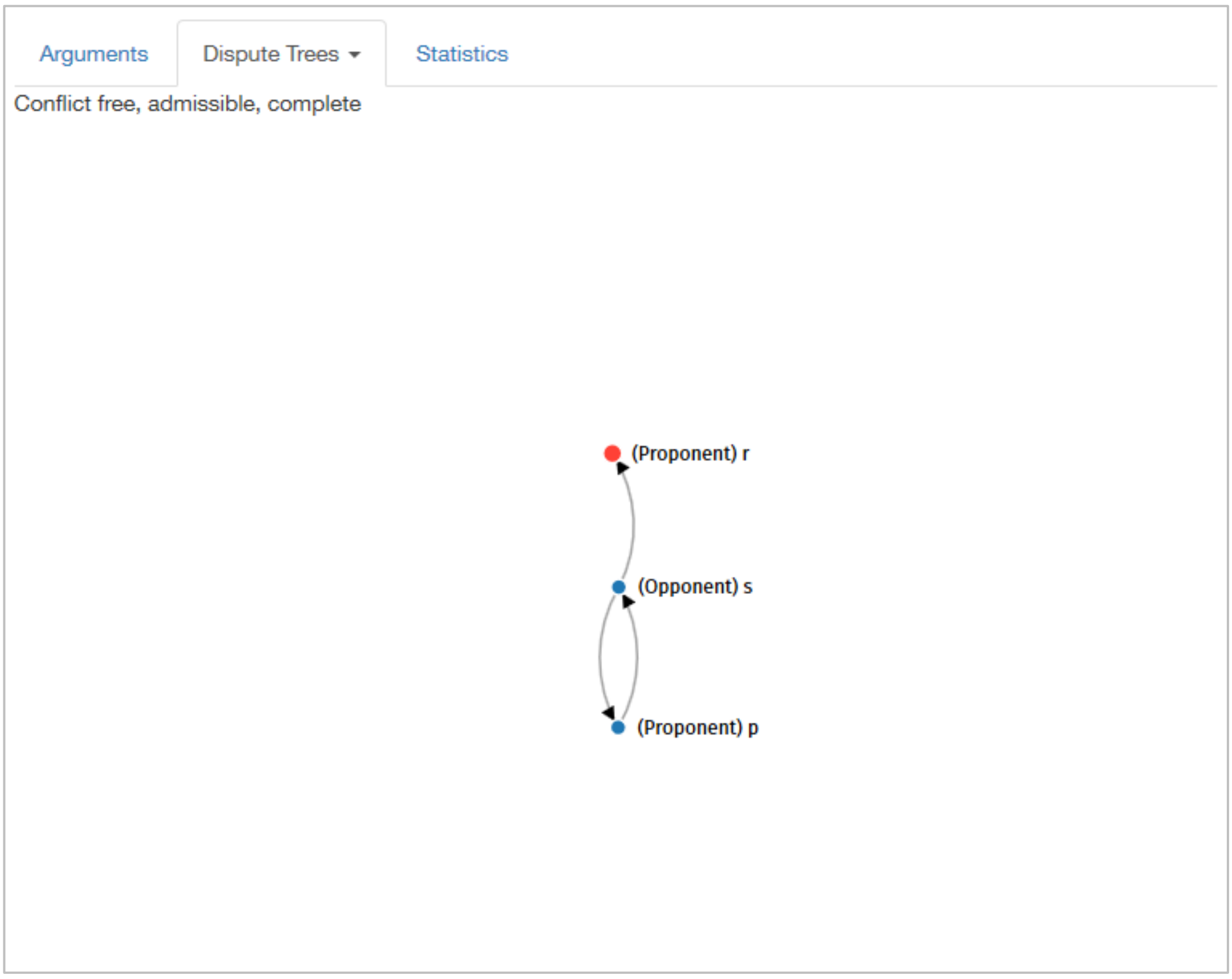

**Figure 11 Visualization of the dispute tree at project web interface.**

### 4.2.1. Syntax for Input

The syntax for input area at the web interface and for *ABA_Parser* class consists of lines of expressions, where each line can only be a single type of expression. There are three types of expression supported in the project, namely:

- Rule [4]

```
supporting |- claim.
```

Supporting symbols (`supporting`) are comma-separated symbols, meanwhile `claim` can only contain one symbol. A symbol is a string of continuous characters (lowercase alphabets, uppercase alphabets, and numeric digits are allowed). For each rule, every symbol present in the rule must be present in order for that argument to be supported. Meanwhile, if there are multiple rules present to support the same claim, the argument is supported when there are at least one rule is presented. For example, when there are the following rules:

```
a, b, c |- p.
d, e |- p.
f, g |- q.
```

[4] Please note that this syntax is different from the usual notation for defining an argument. In the formal definition, to define a sentence, one need to use the notation "$s_0 \leftarrow s_1, \dots, s_m$"; meanwhile the symbol "$\vdash$" is only used for defining an argument where the left hand side is a set of assumptions supporting the claim symbol at right hand side.

This means that *p* can be supported by either "*a*, *b*, *c*" or "*d*, *e*" or both. Meanwhile, *q* can only be supported by "*f*, *g*". For instance, for the first rule, *a*, *b*, and *c* need to be present in order for the rule to be present.

- Contrary:

```
contrary(a, b).
```

This expression means the assumption *a* can be attacked by argument with *b* on its claim. This also implicitly declares *a* as an assumption.

- Assumption:

```
assumption(a).
```

This expression explicitly states that *a* is an assumption. Every assumption needs to have its contrary. This is useful when declaring an argument with no symbols supporting it, i.e. the root itself is an assumption.

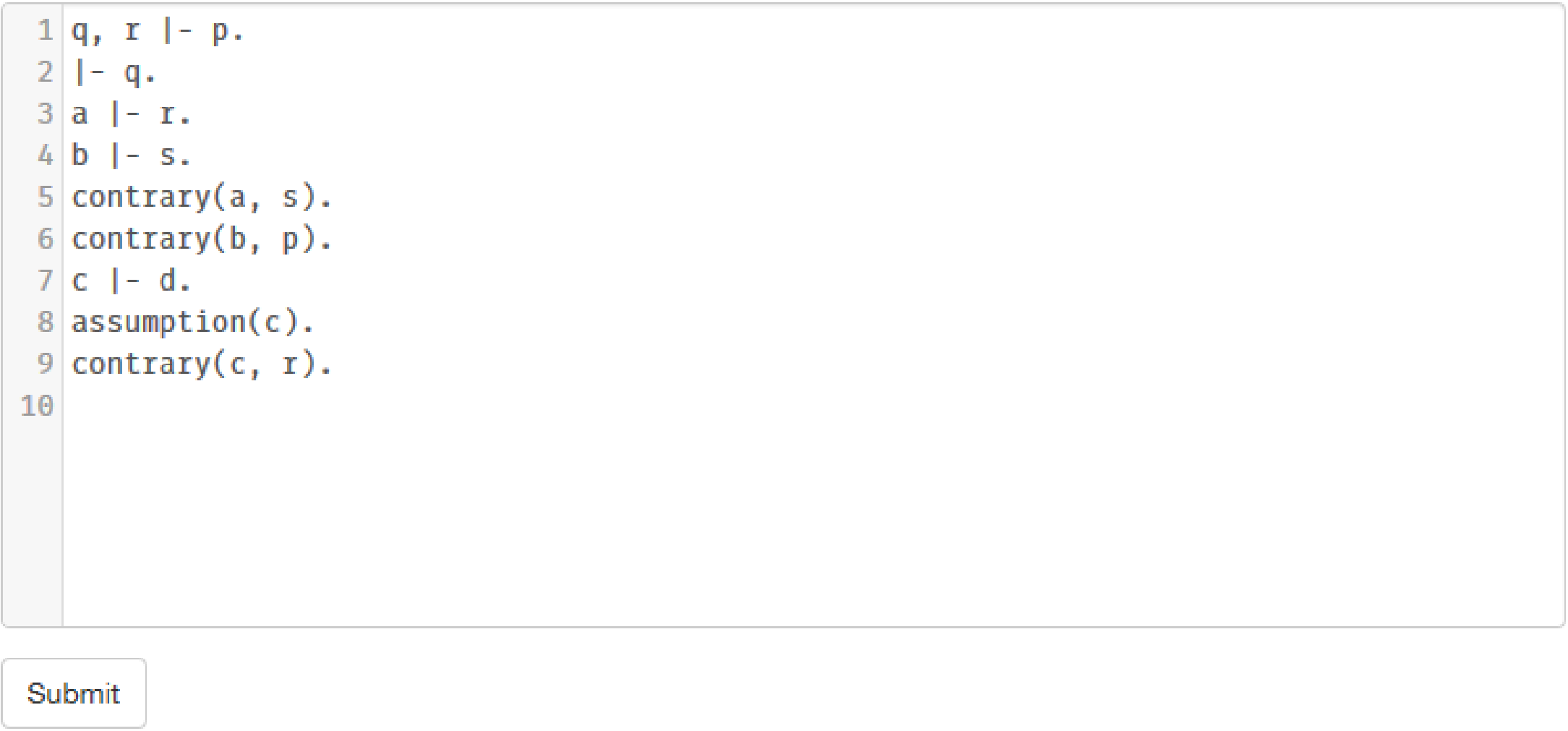

**Figure 12 Input box at aba-web web interface.**

### 4.2.2. Drawing Argument Trees

Argument trees are drawn using D3.js library. The root node of argument is drawn slightly bigger and with a different colour (red) than the rest of the nodes. For each other nodes supporting an argument, the colours used are the same inside the same argument tree. To determine the position of the nodes within the tree, there is an internal variable tracking the node's depth and hence could be used to determine which node shall be drawn at the top or bottom of the tree, so the tree structure is more visible. The edges are drawn from supported nodes to supporting nodes, and since is directional, a black dot is drawn near the supporting nodes (i.e. symbols that are at LHS of the inference rules).

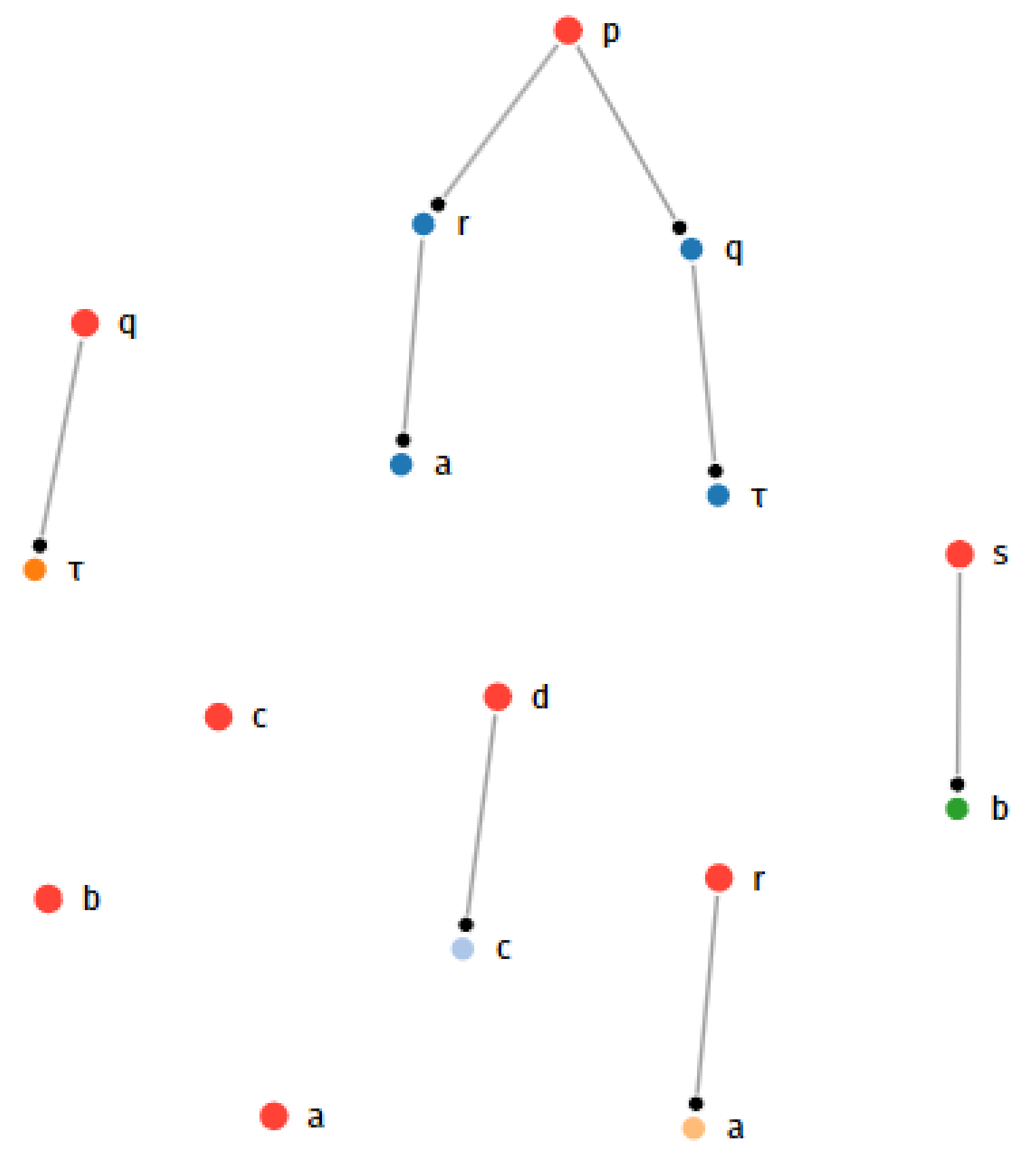

**Figure 13 Argument trees drawn at aba-web's web interface.**

### 4.2.3. Drawing Dispute Trees

Dispute trees are drawn using D3.js library, similar to drawing argument trees, but the edges are specified to have an arrowhead, to represent attacks. The edges are drawn in a curvy arc because if not, the arrows will overlap to other edges and will not be visible to the user.

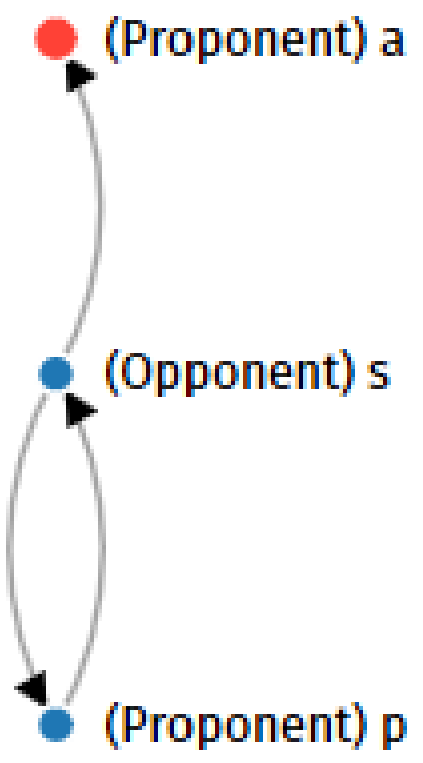

**Figure 14 Dispute tree drawn at aba-web's web interface.**

### 4.2.4. Others

One of the innovations present at front-end is lazy rendering technique, where the visualization is rendered only when one request to see it. After switching to other elements, the old visualization rendering is removed, freeing up the computer's memory used for rendering the elements.

Some simple statistics containing several metrics of program run is being collected and presented to the user. The statistics include:

1. Number of symbols
2. Number of potential arguments
3. Number of actual arguments
4. Number of assumptions
5. Number of arguments satisfying the following semantics:
    5.1. Conflict-free
    5.2. Admissible
    5.3. Complete

5.4. Grounded

5.5. Ideal

5.6. Stable

6. Overall runtime, as perceived by user (wall time)
7. Process runtime, as perceived by server’s CPU (CPU time)

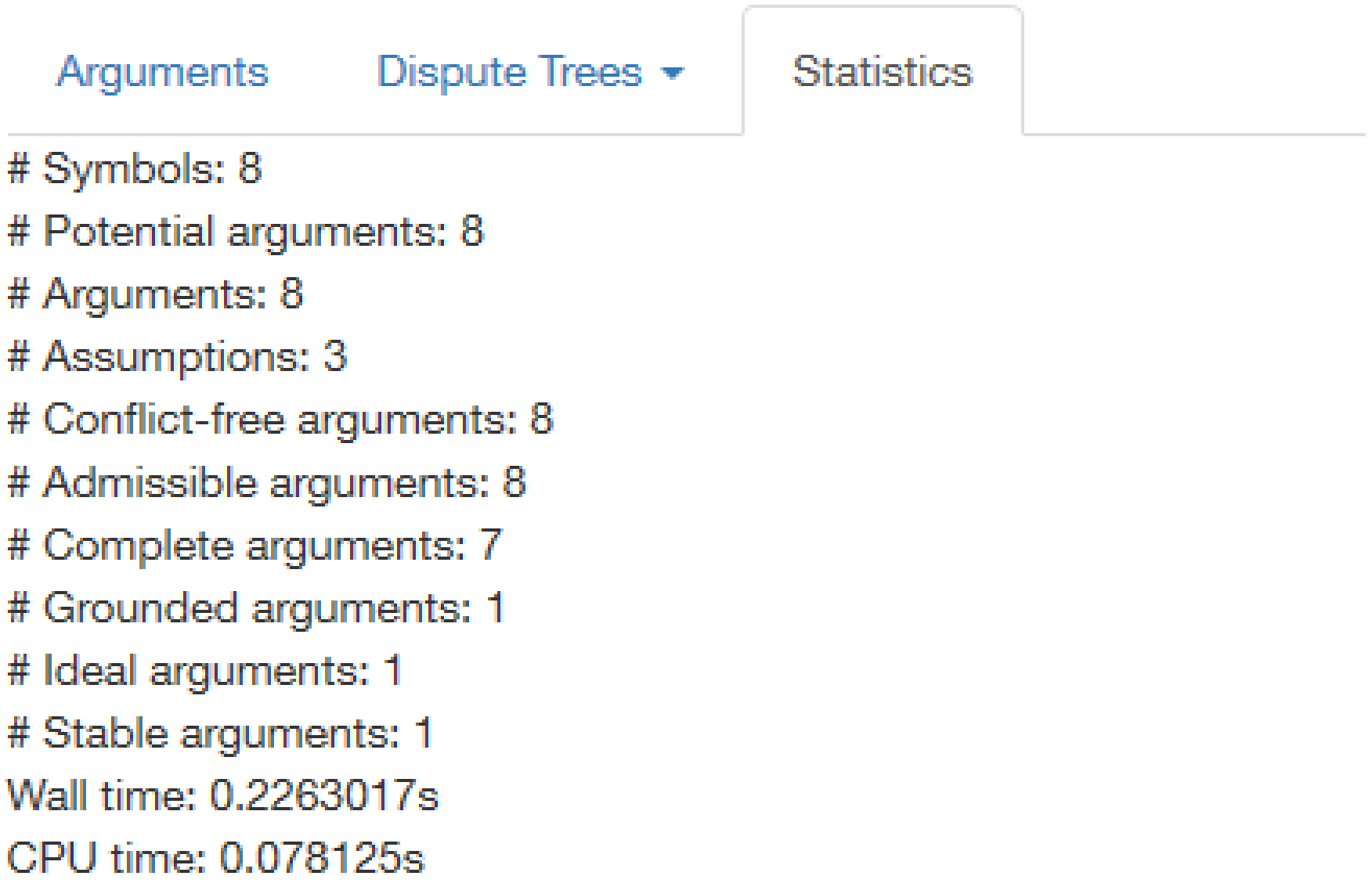

**Figure 15 Statistics of a program run.**

## 4.3. Optimization Techniques

Dispute trees were cached using a simple key-value dictionary during the tree generation phase. The process of storing to cache and loading from cache were shown at Procedure 2.3. and 2.4. respectively. The performance boost of this technique is discussed at Chapter 5.

Dispute tree generation was stopped when there is a tree that satisfies grounded semantics are found. Since if a grounded dispute tree is found, in practice, there is no need to find other dispute trees anymore since the value of the argument semantics (grounded) will not change anymore.

Minor performance speedup was also observed when debugging output is disabled. This was due to the reduction of the need to write output to disk, which may be the bottleneck, depending on the speed of the disk I/O.

Speedup was also observed when changing the method of copying subtrees/branches of a dispute tree during branching. Traditionally, copying an object from one variable to another variable is done by replication of the original pointer to the destination variable. Hence, when a change in the destination variable is observed, the change will be reflected on the original variable. Most of the times, when copying an object, this is not what the user wanted. Hence, Python has a “deep copy” method from its standard library “copy”, where it traverses the variable fields and naively replicates the field and values from one variable to another. Meanwhile, Python has another engineering technique that results in the same effect: the “pickle” technique. This “pickle” technique is implemented by calling Python’s standard “pickle” library, where it dumps a Python object into a binary form in which this can be easily written into disk. But instead writing it to disk, it can be used to deep-copy an object. The performance evaluation of both methods is discussed at Chapter 5.

## 4.4. Others

### 4.4.1. Unit testing

Test cases were written in Python native `unittest` framework to test the correctness of the program. Besides basic tests to verify the correctness of each component, complex test cases were also implemented to verify the correctness of program as a whole. Some test cases are sourced from examples present in Craven and Toni [30]. Running unit test could be carried out using command prompt, as shown in Figure 16.

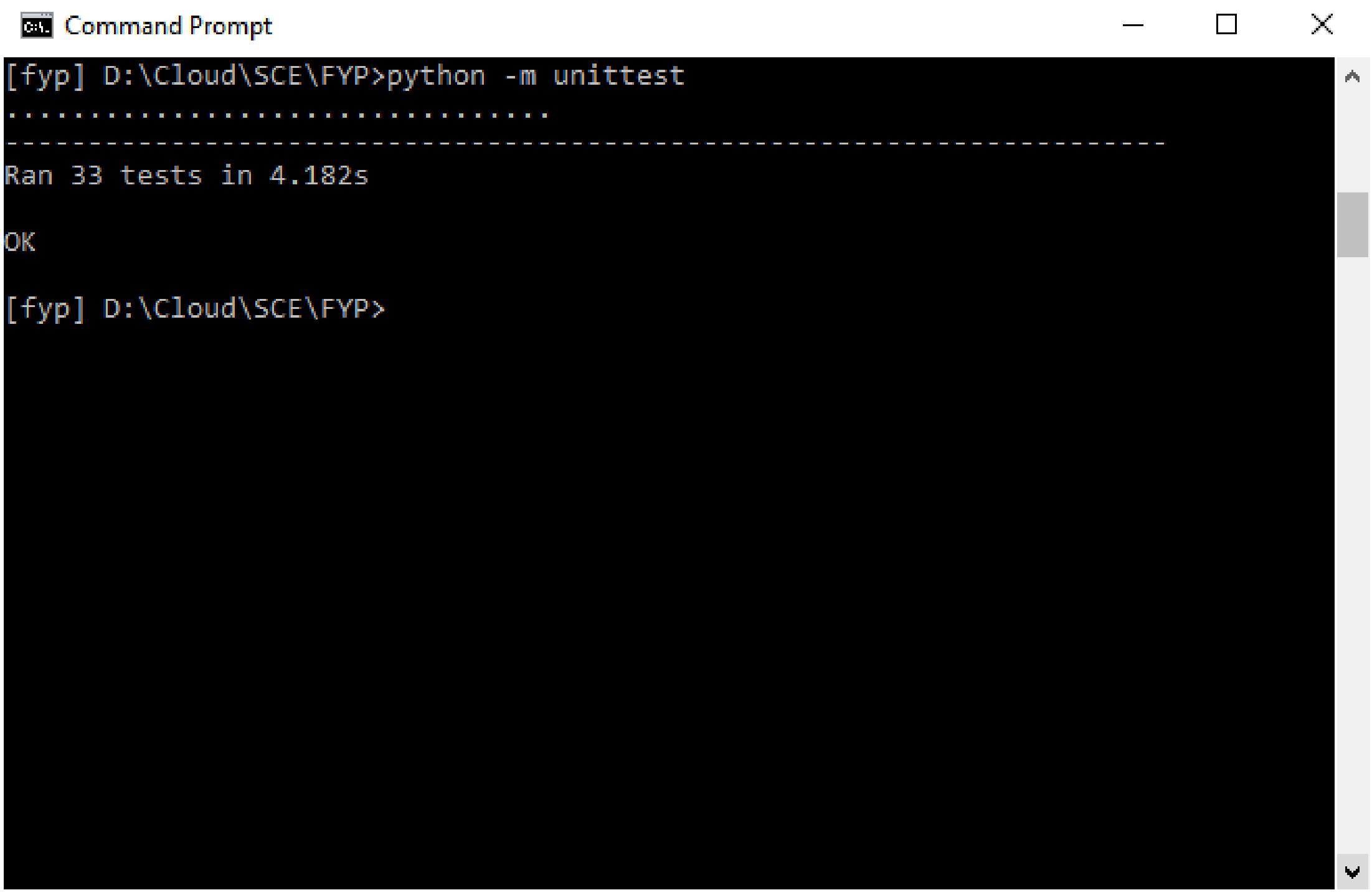

**Figure 16 Running unit tests at local environment.**

To detect if there were any error present when committing the codes, *CircleCI* service was utilized to do the unit testing on the cloud whenever new commits were published to the GitHub repository. Figure 17 shows the list of recent commits being tested and successfully passed the test cases.

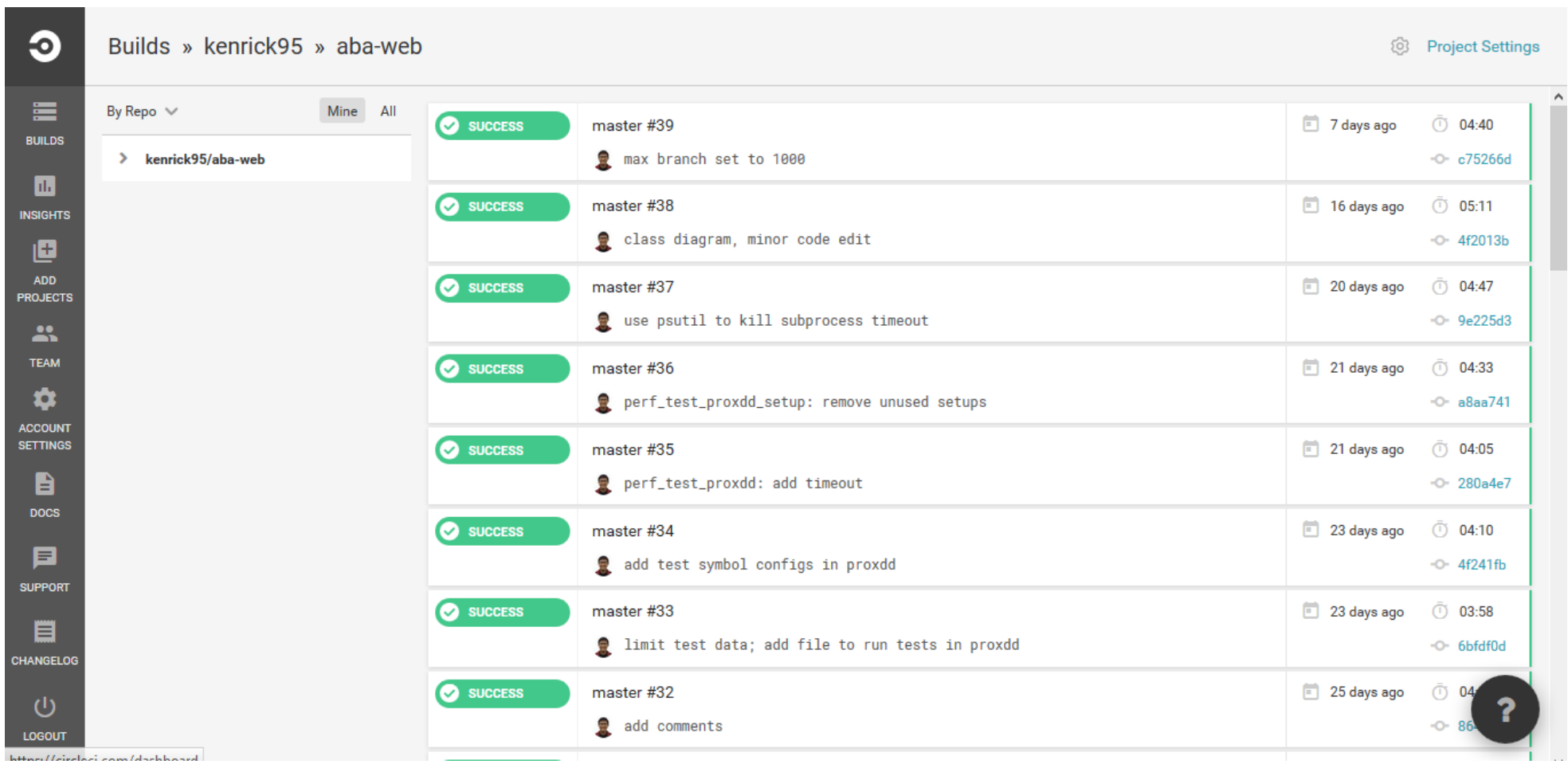

**Figure 17 CircleCI unit testing on every commit pushed to GitHub.**

### 4.4.2. Deployment

After completing the implementation of back-end and front-end, the project was configured, published, and hosted at Microsoft Azure cloud service (dashboard shown at Figure 18). The server was configured to connect to GitHub and the project will be built and updated automatically whenever there are new commits being published to GitHub repository.

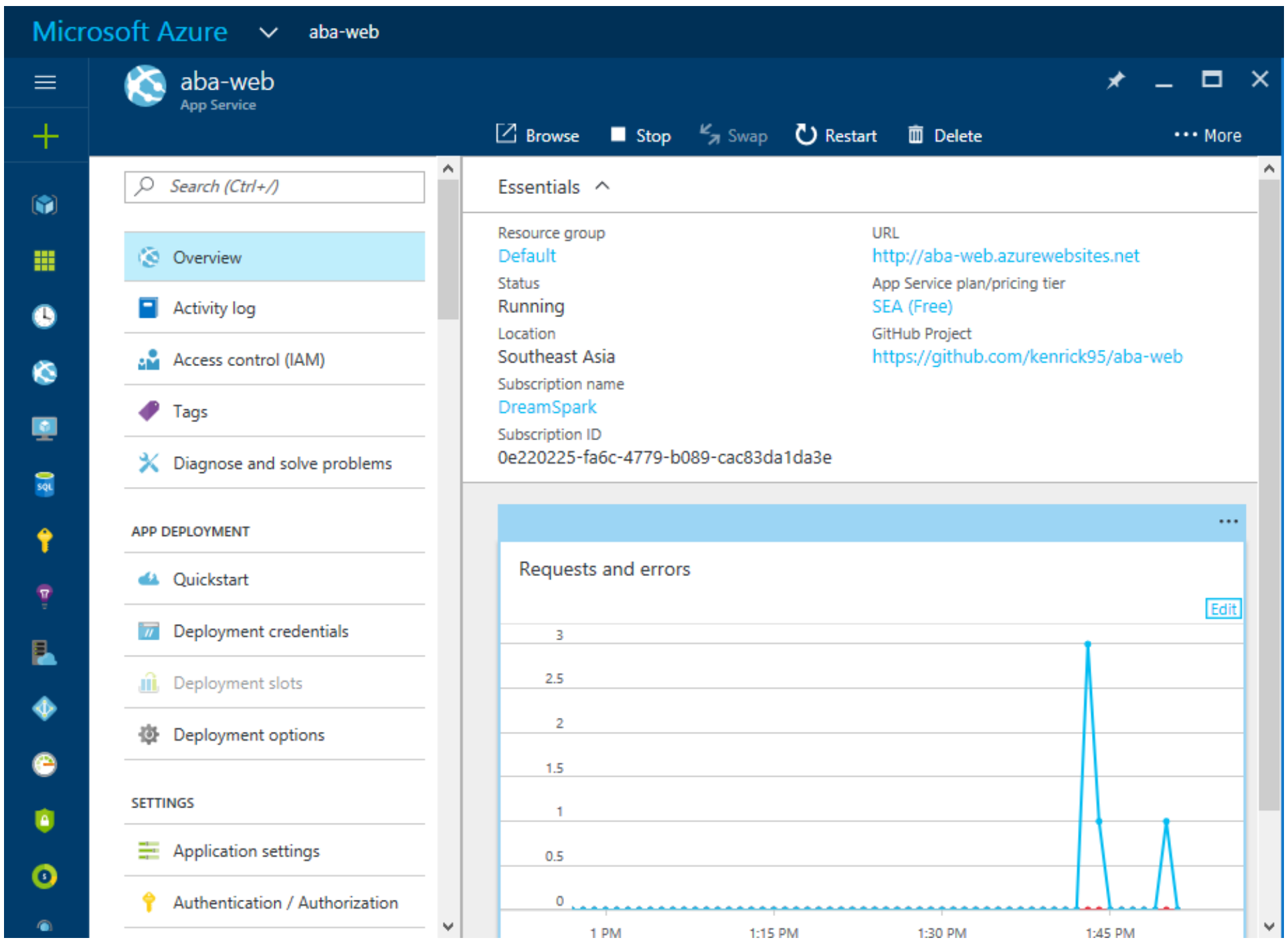

**Figure 18 Microsoft Azure dashboard of *aba-web* project.**

# 5. Experiments

## 5.1. Experiment Design

In evaluating the performance of this project, Toni's Prolog-based *proxdd*[28] was chosen to be compared against because it is the state-of-the-art ABA solver (that computes grounded, admissible, and ideal semantics) and its implementation of solver by representing arguments as trees is similar to the one implemented in this project.

Experiments were conducted on ASUS S46CB notebook computer with Intel Core i7-3537U (2.0 GHz) dual-core processor and 12 GB RAM, running 64-bit Windows 10 Home (Anniversary Update, version 1607, build 14393.321). Python version used is 3.4.5 and Prolog program used is SICStus Prolog 4.3.3. *proxdd* was wrapped in a Python program to record and limit the execution time.

Throughout this chapter, this project will be referred as *aba-web*. Few configurations of *aba-web* were also being compared to answer the following questions:

- Which object copying method is better when branching argument/dispute tree: "pickle" or traditional "copy"?
- What is the effect of dispute tree caching on overall performance?

In this experiment, when referring *aba-web* with "pickle", that means that the "pickle" technique was used; and when referring *aba-web* without "pickle", that means that traditional "copy" method was used.

In comparing among these configurations of *aba-web* and *proxdd*, information that are collected includes:

- the number of successful runs,
- the number of program termination due to runtime error,
- the number of program termination due to memory limit exceeded,
- the number of program termination due to time limit exceeded (timeout error), and
- the average time of successful run.

Time measured was total time taken, as perceived by human, for the program to run from the start until the end, i.e. wall time or wall-clock time; and not the total time the program is being processed in the CPU, i.e. CPU time. Although wall-clock time will be influenced by other programs running in the background, it was being used in the experiment because of technical limitation of not being able to measure the CPU time of *proxdd*, since it was run on another separate process that is being wrapped under a Python program.

In some cases, the experiment could exit due to exceeding memory limit or time limit set by the wrapper program. For *aba-web*, memory limit was arbitrarily set to 5 GB. For *proxdd*, the memory limit was managed by SICStus Prolog (unknown amount). For both program, the time limit was set to 120 seconds.

Test cases used were obtained from Craven and Toni [30], to have a fair comparison with their experiments. They generated the test cases using their Prolog program and published them at Craven's personal website.[5] There were 680 experiment cases in total that they used in their paper. The test cases varied in terms of presence of cyclic arguments, the number of arguments, the number of symbols, the number of assumptions, etc. Ninety-six (96) test cases are selected from the 680 test cases due to the time constraint of running the whole experiment in this project because they reported running the experiments for more than 1,500 hours (nearly 2 months). Since the test cases are randomly generated, it can be said that the 96 test cases represent the 680 test cases available.

At *proxdd*, experiments were conducted individually for each symbol of the test case. After that, the metrics collected were aggregated for each test case. Meanwhile, *aba-web* run the semantics computation for all symbols in one run, hence metrics produced were already aggregated.

[5] http://robertcraven.org/proarg/experiments.html

## 5.2. Results

Different configurations of *aba-web* were compared against Toni's *proxdd*. After running the experiments, taking several hours for each experiment, the results collected were analysed in this section. It should be noted that multiple exceptions could occur during a single run.

Detailed experiment results summary and results per test cases can be found in the Appendix B and Appendix C respectively.

Figure 19 shows the number cases that raised exception across experiments.

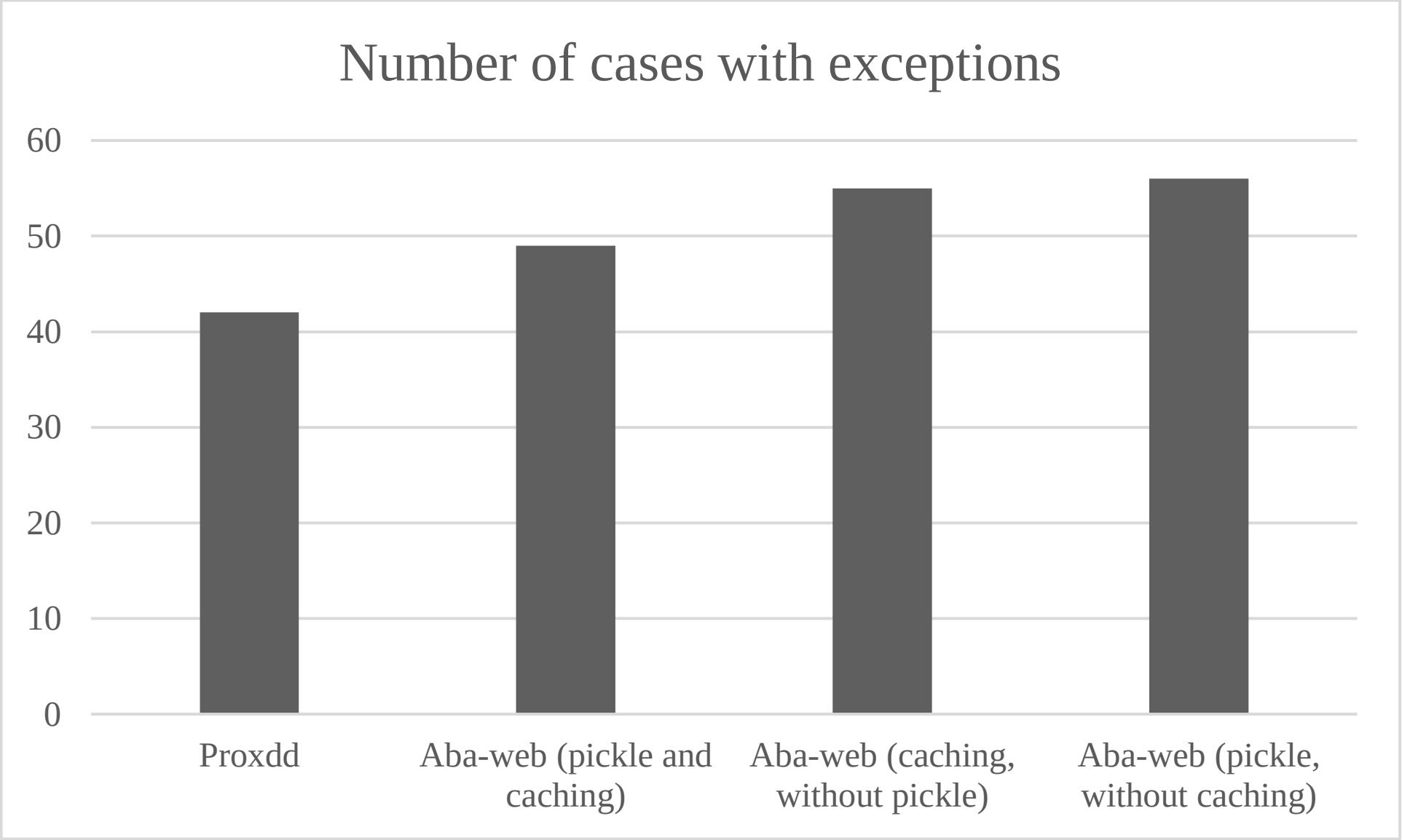

**Figure 19 Total number of cases with exceptions across experiments**

From the data shown in Figure 19, all *aba-web* experiments had more cases with exceptions (49, 55, and 56 cases) compared to that of *proxdd* (42 cases). This might be caused by *aba-web* computing more semantics compared to *proxdd* and hence requiring more resources, and because of that, *aba-web* is more prone to raise an exception when resource limit is exceeded. Meanwhile, for *aba-web*, enabling "pickle" and caching technique minimized the number of cases with exception. Hence, enabling "pickle" and caching technique has a positive effect on the *aba-web*'s overall performance.

Figure 20 shows the number of cases across the experiments that raised runtime exception. This experiment could not detect whether *proxdd* has raised runtime exception due to technical limitation; and if runtime exception has occurred, it will be reported as timeout exception instead. Meanwhile, runtime exceptions in *aba-web* were caught when tree branching was cut due to the tree branching limit (set to 1,000 branches), or when Python's internal maximum recursion depth is reached (using its default value: 1,000).

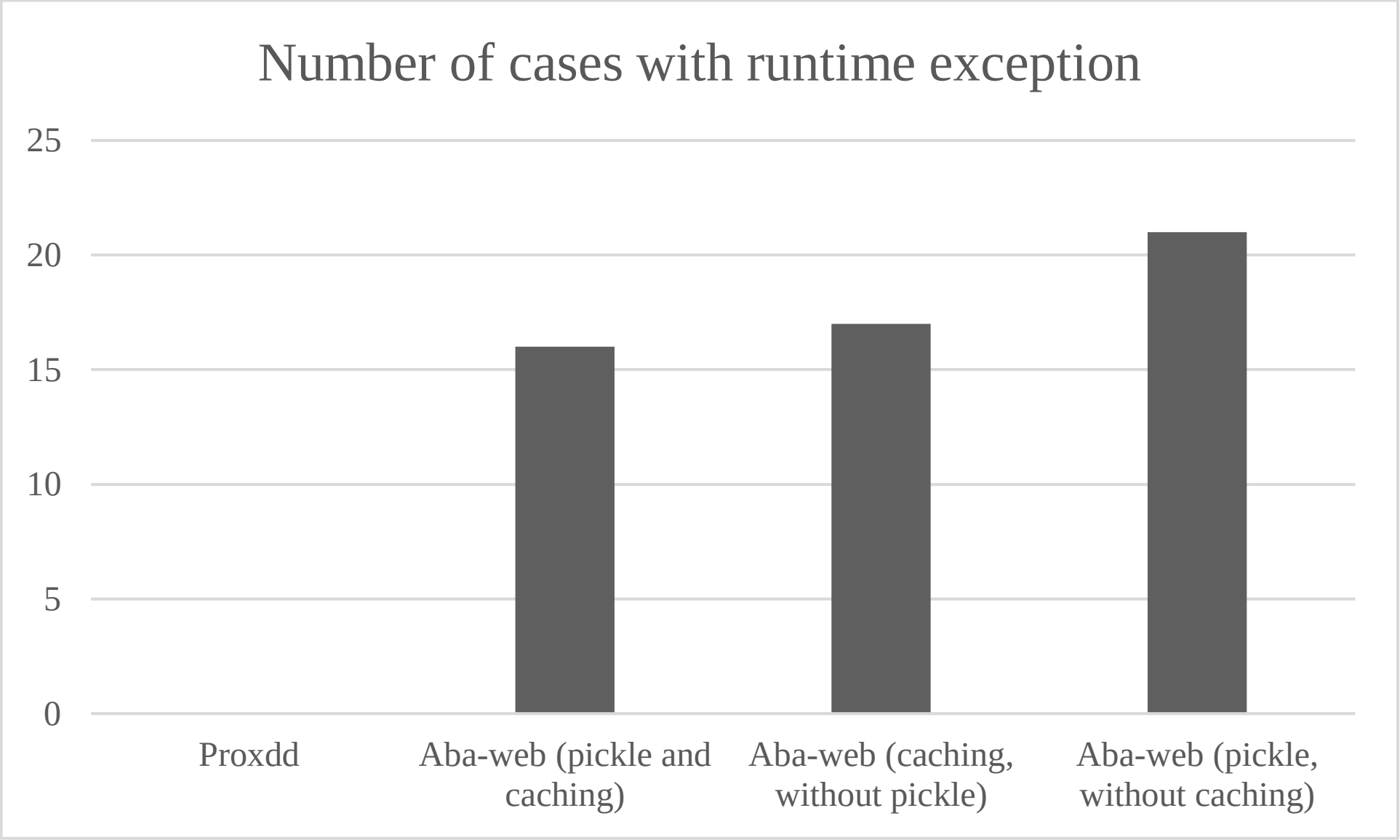

**Figure 20 Total number of cases with runtime exception across experiments**

From Figure 20, since this experiment could not detect whether *proxdd* has reported runtime exception or not, this section only compares the different configurations of *aba-web*. Comparing the different configurations of *aba-web*, it was shown that using "pickle" and caching technique yielded the lowest number of cases with runtime exceptions (16 cases). Usage of caching technique showed that the number of cases with runtime exceptions dropped from 21 to 16 (23.08% reduction). This might be caused by cache hit when enabling the caching technique, because without it, continuing recursive tree expansion might raise Python's maximum recursion depth exception.

Figure 21 shows the number of cases across the experiments that raised timeout exception. The time limit set for each case is 120 seconds.

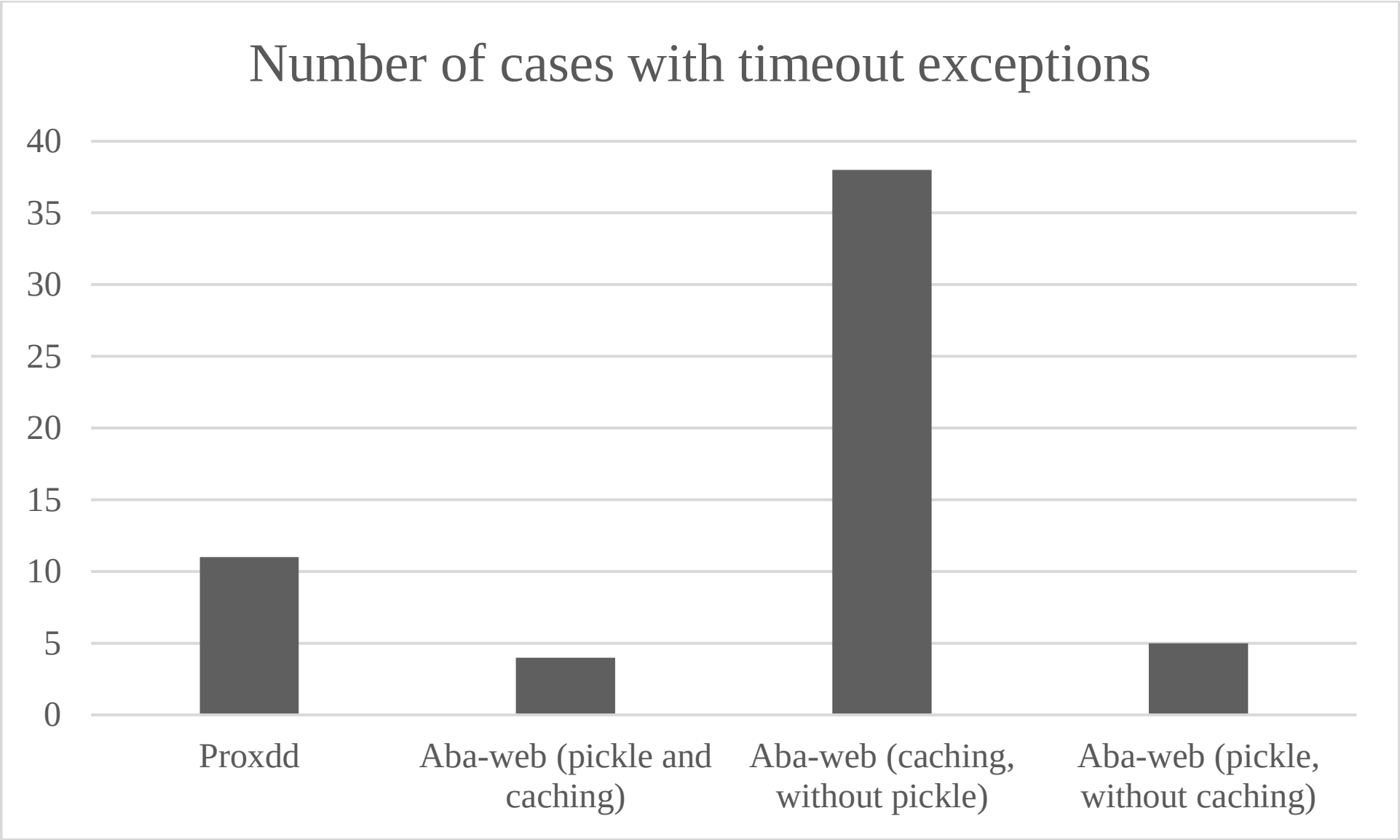

**Figure 21 Total number of cases with timeout exception across experiments**

From Figure 21, it is shown that using "pickle" technique, *aba-web* has significantly lower number of cases with timeout exceptions (4 cases) compared to itself not using "pickle" (38 cases), which shows the effectiveness of the "pickle" technique in speeding up the computations. With tree caching, *aba-web* performed slightly better (4 cases) compared to itself without tree caching (5 cases), hence tree caching only improves the program performance by a little amount, which might be due to a low number of cache hit. Meanwhile, *aba-web* with pickle performs better (4 cases) compared to *proxdd* (11 cases with timeout exceptions), probably because runtime exceptions are combined into timeout exceptions.

Figure 22 shows the number of cases across the experiments that raised memory limit exceeded exception. The memory limit set for *aba-web* experiments is 5 GB, and for *proxdd* is the memory limit set by SICStus Prolog program.

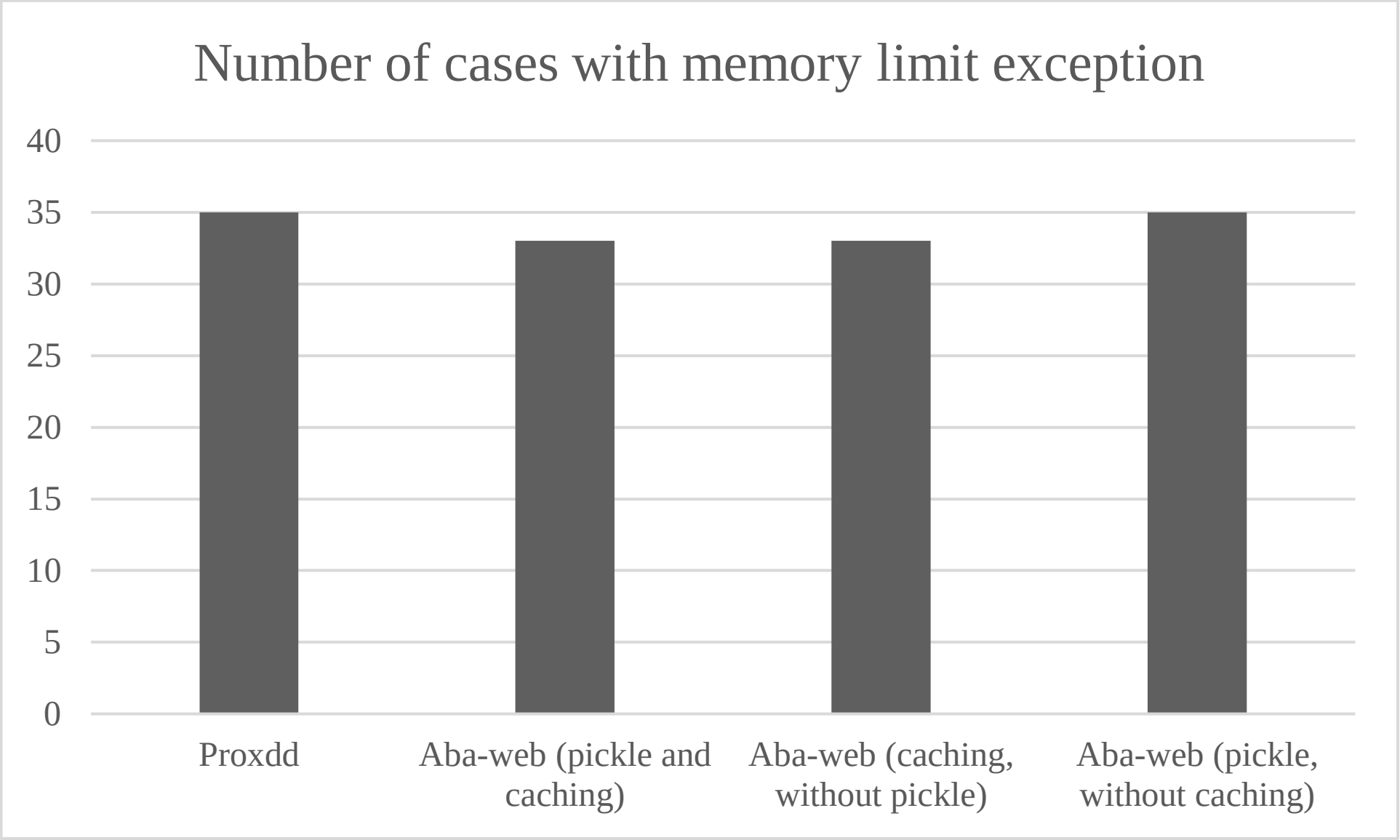

**Figure 22 Total number of cases with memory limit exception across experiments**

From the data in Figure 22, the number of cases with memory limit exceptions was nearly the same across these programs, ranging from 33 to 35 cases. This result was unexpected, especially in *aba-web* with caching enabled, because theoretically, implementing cache causes more memory being used and hence more prone to memory limit exception. This might be caused by the caching technique not significantly affect the memory consumption, since majority of memory consumption came from elsewhere. This might be true, since without caching, the tree recursion will go deeper and hence more prone to runtime or timeout exceptions.

Figure 23 shows the total runtime of all 96 cases across experiments, including those when the programs run for very long but died due to exceptions being raised.

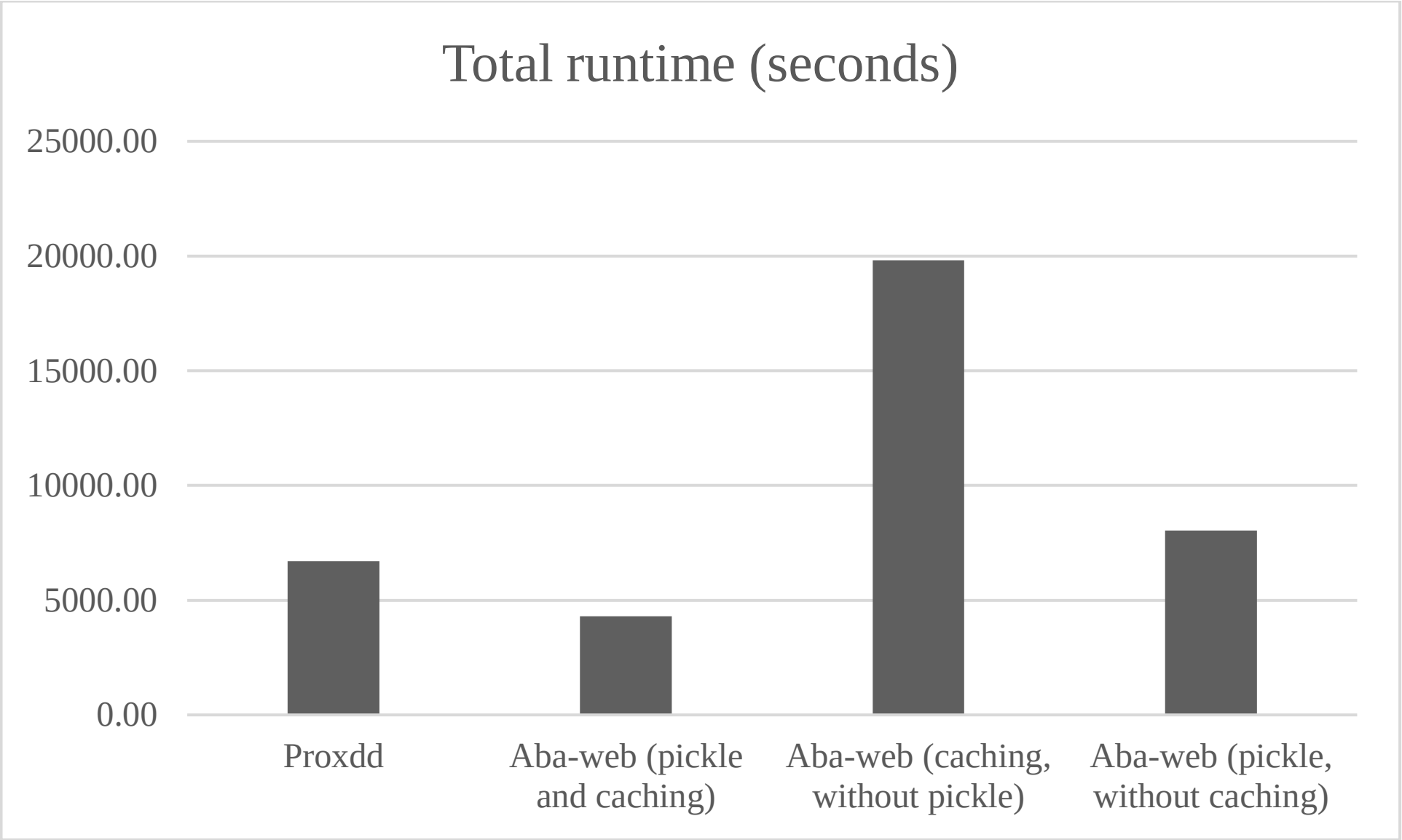

**Figure 23 Total runtime in seconds across experiments**

From results shown in Figure 23, it is observed that the total running time of the program differs a lot, where *aba-web* (with "pickle" technique & caching technique enabled) performed faster than *proxdd* (requiring 35.70% less runtime). Across *aba-web* configurations, it is shown that *aba-web*, with "pickle" and caching enabled, performed the best compared to that when "pickle" technique was disabled (78.27% reduction of runtime) and when caching is disabled (46.51% reduction of runtime).

In the experiments of *aba-web*, the program wasn't directly killed when the time limit has been exceeded, hence the long total runtime, especially in *aba-web* without "pickle", where there were the most number of cases with timeout exceptions. To compare the performance of these programs, the cases that will produce exception need to be ignored, which was done in Figure 24.

Figure 24 shows the average runtime of cases across experiments which did not produce any exceptions during the runtime.

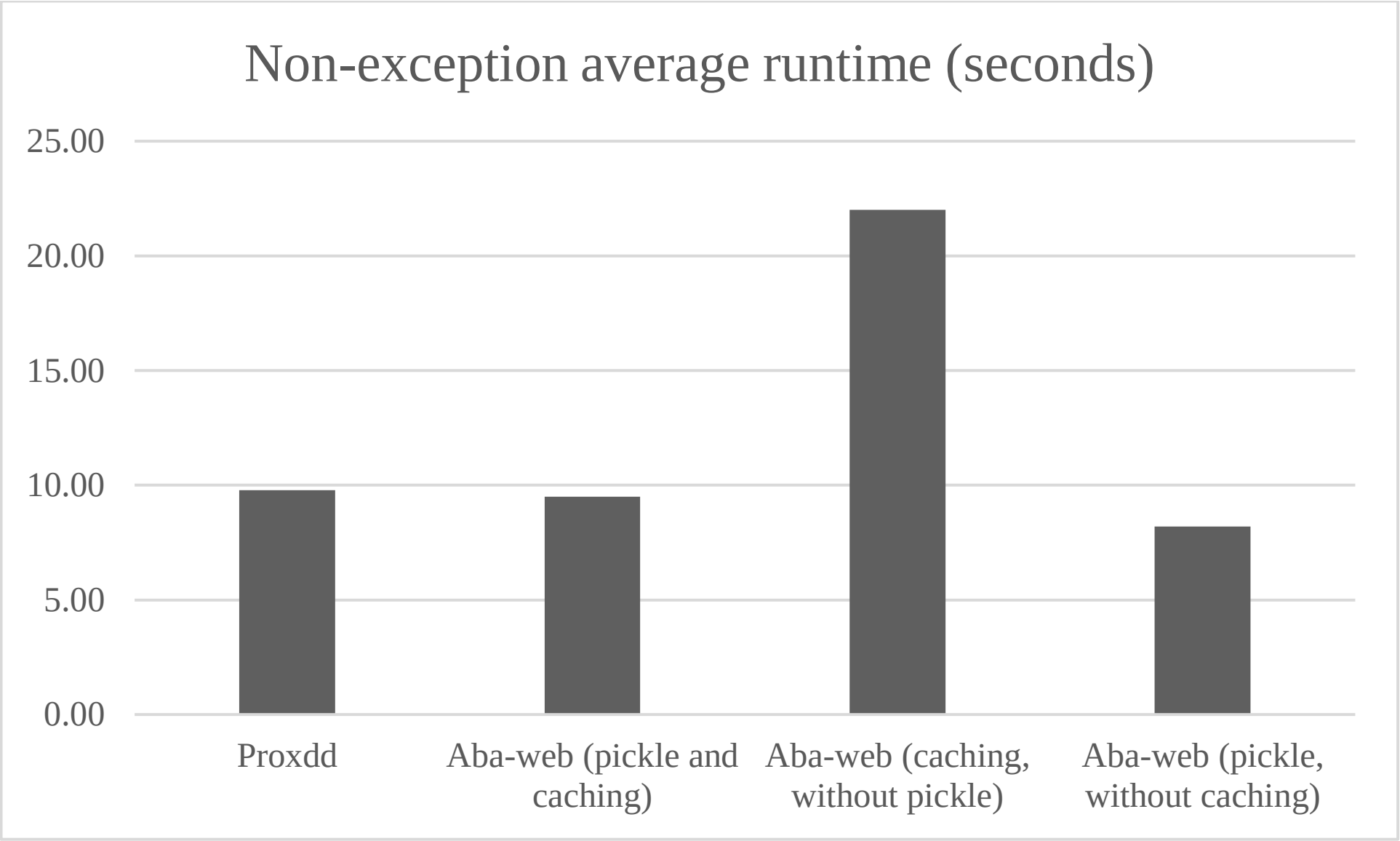

**Figure 24 Average total runtime of cases without exceptions, in seconds, across experiments.**

From results shown in Figure 24, *aba-web* (pickle and caching), and *aba-web* (pickle but without caching) on average run slightly (2.9% and 16.24%, respectively) faster than *proxdd.*

On *aba-web* itself, it is shown that "pickle" technique reduces the average runtime by 56.83% (from 22.01 seconds to 12.51 seconds). This shows that using "pickle" technique is much faster than without it (i.e. traditional "copy" technique). Meanwhile, caching technique unexpectedly increased the average runtime by a small amount 15.85% (from 8.2 seconds to 9.5 seconds). This might be caused by the overhead caused by constructing the cache while not being used in the end.

From the results presented, it can be said that *aba-web* run faster to *proxdd* in terms of average, but slightly inferior in terms of number of cases with exceptions produced. Hence it can be concluded that *aba-web* can run almost as well as the state-of-the-art ABA solver *proxdd*.

# 6. Conclusions

The objective of this project is to implement an Assumption-Based Argumentation (ABA) solver in a modern programming language that performs well to users and create a web interface of the ABA solver so it can be accessible by the public.

This project has demonstrated the novelty of development of an ABA solver in Python with a simple-to-use web interface. At the web interface, the user can describe an ABA framework using a Prolog-like syntax to define inference rules, assumptions, and contraries. Visualization of the arguments and dispute trees are also available in the web interface. Besides that, the project can also be used as a standalone package via command line interface, where it can be interfaced to other projects.

The report has introduced and explained the ABA concepts, such as arguments, attacks, argument trees, dispute trees, and its semantics. The report has also discussed the methodology of this project, laying out the software requirements and software architecture of the project.

The project implemented ABA solver that constructs argument trees, dispute trees, and computes several semantics, such as conflict-free, stable, admissible, grounded, ideal, and complete. The implementation specifics of the project, including back-end, front-end, optimiazation techniques, unit testing, and deployment, were also discussed in the report.

Experiments were designed and conducted in such a way that this project can be compared to *proxdd*, a state-of-the-art tree-based ABA solver by Toni [28], by sourcing and sampling the test cases from Craven and Toni [30] where those test cases were used by them to compare their new approach against *proxdd*.

From the experiment results, in the best configurations of this project, this project run slightly superior to *proxdd* in terms of average runtime of cases where no exceptions were raised (running 3.05% faster than *proxdd*). On the other hand, this project seemed to run somewhat inferior compared to *proxdd* in terms of number of cases with exceptions, where exceptions occurred in 16.67% more cases than *proxdd*. These observations might be because this project computes more semantics and because of that, requires more resources to accomplish these tasks, and hence this project is more prone to exceptions being raised when resource limit was encountered.

Meanwhile, to determine the best project configurations, there were few configurations that were experimented. These experiments were conducted to determine whether to use “pickle” or traditional “copy” method and whether tree caching has an impact on the overall performance of the program.

From the results of the experiments, it can be clearly seen that this project’s best configuration is by employing “pickle” technique (compared to using the traditional “copy” method) and tree caching technique. This configuration yielded the least number of cases with exceptions, where it occurred in 49 out of the 96 experiment cases (51.04%). Besides that, this configuration also produced the least number of cases with memory, timeout, and runtime exceptions (occurred in 33, 4, and 16 out of the 96 experiment cases, respectively). The only trade-off seen was that the configuration only run slower than when tree caching is disabled, where this configuration took 13.65% more time to run, in terms of average time taken for cases without any exceptions.

There were several future works that can be worked on after this project has been completed. Firstly, the scope of this project did not cover the computational complexity analyses as they might have been analysed by other researchers, but analysing the procedures applied in this project will be beneficial to improve its efficiency. After improving the performance of the performance by implementing the algorithms that are the most efficient, this project could be considered to enter the International Competition on Computational Model of Argumentation (ICCMA) to be evaluated against all top-performing argumentation solver.

Besides that, it may be good to implement the other semantics that has been defined over the years by the researchers in argumentation field, for example, “preferred” semantics. This will, in turn, make this project become a universal ABA solver that can be reused and extended in researches in the field of argumentation.

Moreover, a usability testing on the web interface might uncover usability issues so the web interface's user-friendliness could be improved even to users that did not have prior knowledge of argumentation field. This might include an interactive introduction, tutorial, and help page on educating the first-time user to utilize the ABA solver.

Finally, this project could be investigated for usage on multiple disciplines. For example, in medical field, doctors could be trained to utilize an ABA solver in this simple syntax instead of operating traditional solver program in command line. Furthermore, this project could be assessed for commercial usage since commercialization is one way quickly to gain public interest in the field. This can be done by generating more real-life application and impact of this project.

By building an ABA solver in a modern programming language with a web interface[6], the field of argumentation is expected to gain more public interest in turn applied to more real-life scenarios.

[6] The web interface of this project is available on the URL: https://aba-web.azurewebsites.net/

# 7. References

[1] S. Modgil, F. Toni, F. Bex, I. Bratko, C. I. Chesnevar, W. Dvořák*, et al.*, "The added value of argumentation," in *Agreement technologies*, ed: Springer Netherlands, 2013, pp. 357-403.

[2] J. L. Pollock, "Defeasible reasoning," *Cognitive science,* vol. 11, pp. 481-518, 1987.

[3] P. M. Dung, "On the acceptability of arguments and its fundamental role in nonmonotonic reasoning, logic programming and n-person games," *Artificial intelligence,* vol. 77, pp. 321-357, 1995.

[4] A. Bondarenko, P. M. Dung, R. A. Kowalski, and F. Toni, "An abstract, argumentation-theoretic approach to default reasoning," *Artificial intelligence,* vol. 93, pp. 63-101, 1997.

[5] F. Toni, "A tutorial on assumption-based argumentation," *Argument & Computation,* vol. 5, pp. 89-117, 2014.

[6] X. Fan, R. Craven, R. Singer, F. Toni, and M. Williams, "Assumption-based argumentation for decision-making with preferences: A medical case study," in *International Workshop on Computational Logic in Multi-Agent Systems*, 2013, pp. 374-390.

[7] X. Fan and F. Toni, "Decision making with assumption-based argumentation," in *International Workshop on Theorie and Applications of Formal Argumentation*, 2013, pp. 127-142.

[8] L. Carstens, X. Fan, Y. Gao, and F. Toni, "An overview of argumentation frameworks for decision support," in *International Workshop on Graph Structures for Knowledge Representation and Reasoning*, 2015, pp. 32-49.

[9] X. Fan, "Agent dialogues and argumentation," in *The 10th International Conference on Autonomous Agents and Multiagent Systems-Volume 3*, 2011, pp. 1341-1342.

[10] X. Fan and F. Toni, "Assumption-based argumentation dialogues," in *IJCAI*, 2011, pp. 198-203.

[11] X. Fan and F. Toni, "Agent Strategies for ABA-based Information-seeking and Inquiry Dialogues," in *ECAI*, 2012, pp. 324-329.

[12] X. Fan, F. Toni, A. Mocanu, and M. Williams, "Dialogical two-agent decision making with assumption-based argumentation," in *Proceedings of the 2014 international conference on Autonomous agents and multi-agent systems*, 2014, pp. 533-540.

[13] P.-A. Matt, F. Toni, T. Stournaras, and D. Dimitrelos, "Argumentation-based agents for eprocurement," in *Proceedings of the 7th international joint conference on Autonomous agents and multiagent systems: industrial track*, 2008, pp. 71-74.

[14] X. Fan and F. Toni, "On the Interplay between Games, Argumentation and Dialogues," in *Proceedings of the 2016 International Conference on Autonomous Agents & Multiagent Systems*, 2016, pp. 260-268.

[15] F. Toni, "Reasoning on the web with assumption-based argumentation," in *Reasoning Web International Summer School*, 2012, pp. 370-386.

[16] X. Fan and F. Toni, "On Computing Explanations in Argumentation," in *AAAI*, 2015, pp. 1496-1502.

[17] X. Fan and F. Toni, "On Explanations for Non-Acceptable Arguments," in *International Workshop on Theorie and Applications of Formal Argumentation*, 2015, pp. 112-127.

[18] X. Fan and F. Toni, "On Computing Explanations in Abstract Argumentation," in *ECAI*, 2014, pp. 1005-1006.

[19] Q. Zhong, X. Fan, F. Toni, and X. Luo, "Explaining Best Decisions via Argumentation," in *ECSI*, 2014, pp. 224-237.

[20] X. Fan, S. Liu, H. Zhang, C. Leung, and C. Miao, "Explained Activity Recognition with Computational Assumption-based Argumentation," in *Proc. ECAI*, 2016.

[21] X. Fan, H. Zhang, C. Leung, and C. Miao, "A First Step towards Explained Activity Recognition with Computational Abstract Argumentation," *Proc. IEEE MFI,* 2016.

[22] P. E. Dunne and M. Wooldridge, "Complexity of abstract argumentation," in *Argumentation in Artificial Intelligence*, ed: Springer, 2009, pp. 85-104.

[23] P. E. Dunne, "The computational complexity of ideal semantics," *Artificial Intelligence,* vol. 173, pp. 1559-1591, 2009.

[24] Y. Dimopoulos, B. Nebel, and F. Toni, "On the computational complexity of assumption-based argumentation for default reasoning," *Artificial Intelligence,* vol. 141, pp. 57-78, 2002.

[25] U. Egly, S. A. Gaggl, and S. Woltran, "Aspartix: Implementing argumentation frameworks using answer-set programming," in *International Conference on Logic Programming*, 2008, pp. 734-738.
[26] F. CERUTTI, M. VALLATI, and M. GIACOMIN, "Efficient and Off-The-Shelf Solver: jArgSemSAT," *Computational Models of Argument: Proceedings of COMMA 2016,* vol. 287, p. 465, 2016.
[27] M. Snaith and C. Reed, "TOAST: online ASPIC implementation," 2012.
[28] F. Toni, "A generalised framework for dispute derivations in assumption-based argumentation," *Artificial Intelligence,* vol. 195, pp. 1-43, 2013.
[29] R. Craven, F. Toni, and M. Williams, "Graph-based dispute derivations in assumption-based argumentation," in *International Workshop on Theorie and Applications of Formal Argumentation*, 2013, pp. 46-62.
[30] R. Craven and F. Toni, "Argument graphs and assumption-based argumentation," *Artificial Intelligence,* vol. 233, pp. 1-59, 2016.
[31] (2016). *Stack Overflow Developer Survey Results 2016* Available: http://stackoverflow.com/research/developer-survey-2016#technology
[32] H. Prakken, "An abstract framework for argumentation with structured arguments," *Argument and Computation,* vol. 1, pp. 93-124, 2010.
[33] P. Besnard, A. Garcia, A. Hunter, S. Modgil, H. Prakken, G. Simari*, et al.*, "Introduction to structured argumentation," *Argument & Computation,* vol. 5, pp. 1-4, 2014.
[34] P. M. Dung, R. A. Kowalski, and F. Toni, "Assumption-based argumentation," in *Argumentation in Artificial Intelligence*, ed: Springer, 2009, pp. 199-218.
[35] P. M. Dung, P. Mancarella, and F. Toni, "Computing ideal sceptical argumentation," *Artificial Intelligence,* vol. 171, pp. 642-674, 2007.
[36] P. M. Dung, R. A. Kowalski, and F. Toni, "Dialectic proof procedures for assumption-based, admissible argumentation," *Artificial Intelligence,* vol. 170, pp. 114-159, 2006.
[37] B. Bruegge and A. H. Dutoit, "Object-Oriented Software Engineering," ed: Prentice Hall, 1999.

# 8. Appendices

## Appendix A: Program Listing

Detailed program listing is available on GitHub at the following URL:

https://github.com/kenrick95/aba-web

Web interface of the ABA solver is available at the following URL:

https://aba-web.azurewebsites.net/

## Appendix B: Summary of Experiment Results

**Table 1 Summarized experiment results, comparing performance of *proxdd* and *aba-web* in different configurations**

| | **Proxdd** | **Aba-web (pickle and caching)** | **Aba-web (caching, without pickle)** | **Aba-web (pickle, without caching)** |
|---|---|---|---|---|
| *Number of cases with exceptions* | 42 | 49 | 55 | 56 |
| *Number of cases with timeout exception* | 11 | 4 | 38 | 5 |
| *Number of cases with runtime exception* | 0 | 16 | 17 | 21 |
| *Number of cases with memory limit exception* | 35 | 33 | 33 | 35 |
| *Total runtime* | 6691.82s | 4303.03s | 19802.98s | 8044.75s |
| *Number of cases without exceptions* | 54 | 47 | 41 | 40 |
| *Non-exception total runtime* | 528.40s | 446.27s | 902.41s | 327.97s |
| *Non-exception average runtime* | 9.7852s | 9.4952s | 22.0100s | 8.1992s |

# Appendix C: Detailed Performance Evaluation Results

## Proxdd

| | Wall time (s) | CPU time (s) | Exception |
|---|---|---|---|
| exp_acyclic_depvary_step1_batch_yyy01 | 0 | 8.717462755 | |
| exp_acyclic_depvary_step1_batch_yyy02 | 0.046875 | 12.19307845 | |
| exp_acyclic_depvary_step1_batch_yyy03 | 0.015625 | 8.168678952 | |
| exp_acyclic_depvary_step1_batch_yyy04 | 0 | 8.569151884 | |
| exp_acyclic_depvary_step2_batch_yyy01 | 0.046875 | 8.346992245 | |
| exp_acyclic_depvary_step2_batch_yyy02 | 0.046875 | 9.852543855 | |
| exp_acyclic_depvary_step2_batch_yyy03 | 0.09375 | 8.630505733 | |
| exp_acyclic_depvary_step2_batch_yyy04 | 0.015625 | 8.202013244 | |
| exp_acyclic_depvary_step3_batch_yyy01 | 0.015625 | 8.771516542 | |
| exp_acyclic_depvary_step3_batch_yyy02 | 0.015625 | 8.136476085 | |
| exp_acyclic_depvary_step3_batch_yyy03 | 0.015625 | 7.586231203 | |
| exp_acyclic_depvary_step3_batch_yyy04 | 0.015625 | 7.78426517 | |
| exp_acyclic_indvary1_step1_batch_yyy01 | 0.03125 | 8.218891402 | |
| exp_acyclic_indvary1_step1_batch_yyy02 | 0.015625 | 8.099757787 | |
| exp_acyclic_indvary1_step1_batch_yyy03 | 0.0625 | 9.131568598 | |
| exp_acyclic_indvary1_step1_batch_yyy04 | 0.046875 | 7.992655601 | |
| exp_acyclic_indvary1_step2_batch_yyy01 | 0.28125 | 138.123259 | Timeout |
| exp_acyclic_indvary1_step2_batch_yyy02 | 0.4375 | 247.0993161 | Timeout |
| exp_acyclic_indvary1_step2_batch_yyy03 | 0.453125 | 247.6446793 | Timeout |
| exp_acyclic_indvary1_step2_batch_yyy04 | 0.609375 | 372.9890577 | Timeout |
| exp_acyclic_indvary1_step3_batch_yyy01 | 0.03125 | 7.858438258 | |
| exp_acyclic_indvary1_step3_batch_yyy02 | 0 | 8.601849031 | |
| exp_acyclic_indvary1_step3_batch_yyy03 | 0.03125 | 8.633582252 | |
| exp_acyclic_indvary1_step3_batch_yyy04 | 0 | 15.96179763 | |
| exp_acyclic_indvary2_step1_batch_yyy01 | 0.03125 | 9.897907524 | |
| exp_acyclic_indvary2_step1_batch_yyy02 | 0.0625 | 9.947594083 | |
| exp_acyclic_indvary2_step1_batch_yyy03 | 0.0625 | 9.741281349 | |
| exp_acyclic_indvary2_step1_batch_yyy04 | 0.046875 | 9.270853124 | |
| exp_acyclic_indvary2_step2_batch_yyy01 | 0.203125 | 130.5303177 | Timeout |
| exp_acyclic_indvary2_step2_batch_yyy02 | 0.0625 | 7.665009216 | |
| exp_acyclic_indvary2_step2_batch_yyy03 | 0.046875 | 8.734523599 | |
| exp_acyclic_indvary2_step2_batch_yyy04 | 0.0625 | 13.26822737 | |
| exp_acyclic_indvary2_step3_batch_yyy01 | 0.03125 | 44.56378501 | |
| exp_acyclic_indvary2_step3_batch_yyy02 | 0.5 | 248.0138751 | Timeout |
| exp_acyclic_indvary2_step3_batch_yyy03 | 0.046875 | 15.14618104 | |
| exp_acyclic_indvary2_step3_batch_yyy04 | 0.03125 | 8.512307718 | |
| exp_acyclic_indvary3_step1_batch_yyy01 | 0.03125 | 8.006946594 | |
| exp_acyclic_indvary3_step1_batch_yyy02 | 0.046875 | 8.324383485 | |
| exp_acyclic_indvary3_step1_batch_yyy03 | 0.015625 | 11.31103425 | |
| exp_acyclic_indvary3_step1_batch_yyy04 | 0.015625 | 9.143283919 | |
| exp_acyclic_indvary3_step2_batch_yyy01 | 0.015625 | 8.170720523 | |
| exp_acyclic_indvary3_step2_batch_yyy02 | 0.015625 | 8.327150053 | |
| exp_acyclic_indvary3_step2_batch_yyy03 | 0.015625 | 10.60674215 | |
| exp_acyclic_indvary3_step2_batch_yyy04 | 0.015625 | 8.496664847 | |
| exp_acyclic_indvary3_step3_batch_yyy01 | 0.015625 | 7.860538945 | |
| exp_acyclic_indvary3_step3_batch_yyy02 | 0.03125 | 8.617873286 | |
| exp_acyclic_indvary3_step3_batch_yyy03 | 0.046875 | 11.90138924 | |
| exp_acyclic_indvary3_step3_batch_yyy04 | 0.046875 | 8.181719468 | |
| exp_cycles_depvary_step1_batch_yyy01 | 0.046875 | 8.326565047 | |
| exp_cycles_depvary_step1_batch_yyy02 | 0.046875 | 8.895699688 | |
| exp_cycles_depvary_step1_batch_yyy03 | 0.046875 | 372.9101093 | Memory |
| exp_cycles_depvary_step1_batch_yyy04 | 0.109375 | 10.80053656 | |
| exp_cycles_depvary_step2_batch_yyy01 | 0.015625 | 89.69940719 | Memory |
| exp_cycles_depvary_step2_batch_yyy02 | 0.0625 | 8.37160768 | |
| exp_cycles_depvary_step2_batch_yyy03 | 0 | 8.420186217 | |
| exp_cycles_depvary_step2_batch_yyy04 | 0.046875 | 163.1644415 | Memory |
| exp_cycles_depvary_step3_batch_yyy01 | 0.046875 | 7.85542373 | |
| exp_cycles_depvary_step3_batch_yyy02 | 0.265625 | 278.323856 | Memory & timeout |
| exp_cycles_depvary_step3_batch_yyy03 | 0.015625 | 120.944674 | Memory |
| exp_cycles_depvary_step3_batch_yyy04 | 0.171875 | 200.9195515 | Memory & timeout |
| exp_cycles_indvary1_step1_batch_yyy01 | 0.0625 | 171.2657461 | Memory |
| exp_cycles_indvary1_step1_batch_yyy02 | 0.015625 | 7.794103134 | |
| exp_cycles_indvary1_step1_batch_yyy03 | 0.03125 | 22.67873614 | Memory |
| exp_cycles_indvary1_step1_batch_yyy04 | 0.015625 | 23.71227611 | Memory |
| exp_cycles_indvary1_step2_batch_yyy01 | 0.015625 | 160.088974 | Memory |
| exp_cycles_indvary1_step2_batch_yyy02 | 0.015625 | 130.1264415 | Memory |
| exp_cycles_indvary1_step2_batch_yyy03 | 0.046875 | 81.35193791 | Memory |
| exp_cycles_indvary1_step2_batch_yyy04 | 0.015625 | 50.13301818 | Memory |
| exp_cycles_indvary1_step3_batch_yyy01 | 0 | 100.7704315 | Memory |
| exp_cycles_indvary1_step3_batch_yyy02 | 0 | 106.6274694 | Memory |
| exp_cycles_indvary1_step3_batch_yyy03 | 0.03125 | 113.2693576 | Memory |
| exp_cycles_indvary1_step3_batch_yyy04 | 0.046875 | 75.1178934 | Memory |
| exp_cycles_indvary2_step1_batch_yyy01 | 0.5625 | 415.6599929 | Memory & timeout |
| exp_cycles_indvary2_step1_batch_yyy02 | 0.046875 | 51.63815926 | Memory |
| exp_cycles_indvary2_step1_batch_yyy03 | 0.21875 | 378.0968126 | Memory & timeout |
| exp_cycles_indvary2_step1_batch_yyy04 | 0.03125 | 72.47039374 | Memory |
| exp_cycles_indvary2_step2_batch_yyy01 | 0.015625 | 38.72996753 | Memory |
| exp_cycles_indvary2_step2_batch_yyy02 | 0.03125 | 58.22308087 | Memory |
| exp_cycles_indvary2_step2_batch_yyy03 | 0.015625 | 20.01747589 | Memory |
| exp_cycles_indvary2_step2_batch_yyy04 | 0.046875 | 13.23939865 | |
| exp_cycles_indvary2_step3_batch_yyy01 | 0.03125 | 59.13658775 | Memory |
| exp_cycles_indvary2_step3_batch_yyy02 | 0.015625 | 7.367167788 | |
| exp_cycles_indvary2_step3_batch_yyy03 | 0.03125 | 35.1359231 | Memory |
| exp_cycles_indvary2_step3_batch_yyy04 | 0.015625 | 36.83035753 | Memory |
| exp_cycles_indvary3_step1_batch_yyy01 | 0.015625 | 7.38438874 | |
| exp_cycles_indvary3_step1_batch_yyy02 | 0.015625 | 7.485642912 | |
| exp_cycles_indvary3_step1_batch_yyy03 | 0 | 7.307152681 | |
| exp_cycles_indvary3_step1_batch_yyy04 | 0.25 | 127.1881184 | Timeout |
| exp_cycles_indvary3_step2_batch_yyy01 | 0.046875 | 144.5085054 | Memory |
| exp_cycles_indvary3_step2_batch_yyy02 | 0.015625 | 218.1926696 | Memory |
| exp_cycles_indvary3_step2_batch_yyy03 | 0.015625 | 99.964143 | Memory |
| exp_cycles_indvary3_step2_batch_yyy04 | 0 | 248.7579206 | Memory |
| exp_cycles_indvary3_step3_batch_yyy01 | 0 | 129.962507 | Memory |
| exp_cycles_indvary3_step3_batch_yyy02 | 0.015625 | 189.7426505 | Memory |
| exp_cycles_indvary3_step3_batch_yyy03 | 0.015625 | 40.60401212 | Memory |
| exp_cycles_indvary3_step3_batch_yyy04 | 0 | 153.0573504 | Memory |

Aba-web (Pickle and cache)

| | Wall time (s) | CPU time (s) | Exception |
|---|---|---|---|
| exp_acyclic_depvary_step1_batch_yyy01 | 0.013523705 | 0.015625 | |
| exp_acyclic_depvary_step1_batch_yyy02 | 0.010115068 | 0.015625 | |
| exp_acyclic_depvary_step1_batch_yyy03 | 0.013694075 | 0.015625 | |
| exp_acyclic_depvary_step1_batch_yyy04 | 0.012209596 | 0.015625 | |
| exp_acyclic_depvary_step2_batch_yyy01 | 0.024073524 | 0.015625 | |
| exp_acyclic_depvary_step2_batch_yyy02 | 0.034669734 | 0.03125 | |
| exp_acyclic_depvary_step2_batch_yyy03 | 0.12995721 | 0.140625 | |
| exp_acyclic_depvary_step2_batch_yyy04 | 0.030694154 | 0.03125 | |
| exp_acyclic_depvary_step3_batch_yyy01 | 0.196115476 | 0.1875 | |
| exp_acyclic_depvary_step3_batch_yyy02 | 0.157563763 | 0.171875 | |
| exp_acyclic_depvary_step3_batch_yyy03 | 0.197419732 | 0.1875 | |
| exp_acyclic_depvary_step3_batch_yyy04 | 0.304010352 | 0.3125 | |
| exp_acyclic_indvary1_step1_batch_yyy01 | 9.18278431 | 8.96875 | Runtime |
| exp_acyclic_indvary1_step1_batch_yyy02 | 6.449580048 | 6.359375 | Runtime |
| exp_acyclic_indvary1_step1_batch_yyy03 | 11.56034535 | 11.5625 | Runtime |
| exp_acyclic_indvary1_step1_batch_yyy04 | 37.09550095 | 37.09375 | |
| exp_acyclic_indvary1_step2_batch_yyy01 | 154.562382 | 154.484375 | Memory & timeout |
| exp_acyclic_indvary1_step2_batch_yyy02 | 1144.231452 | 1143.515625 | Memory & timeout |
| exp_acyclic_indvary1_step2_batch_yyy03 | 316.0586287 | 315.765625 | Memory & timeout |
| exp_acyclic_indvary1_step2_batch_yyy04 | 346.6438691 | 346.4375 | Runtime & timeout |
| exp_acyclic_indvary1_step3_batch_yyy01 | 1.829362885 | 1.828125 | Runtime |
| exp_acyclic_indvary1_step3_batch_yyy02 | 0.585583966 | 0.59375 | |
| exp_acyclic_indvary1_step3_batch_yyy03 | 9.28768847 | 9.296875 | |
| exp_acyclic_indvary1_step3_batch_yyy04 | 4.955407327 | 4.953125 | Runtime |
| exp_acyclic_indvary2_step1_batch_yyy01 | 2.058593415 | 2.046875 | |
| exp_acyclic_indvary2_step1_batch_yyy02 | 0.317141179 | 0.328125 | |
| exp_acyclic_indvary2_step1_batch_yyy03 | 9.772487952 | 9.78125 | |
| exp_acyclic_indvary2_step1_batch_yyy04 | 5.149235735 | 5.140625 | Runtime |
| exp_acyclic_indvary2_step2_batch_yyy01 | 52.8785457 | 52.875 | |
| exp_acyclic_indvary2_step2_batch_yyy02 | 76.11856744 | 76.0625 | Memory |
| exp_acyclic_indvary2_step2_batch_yyy03 | 76.69856368 | 76.59375 | Memory |
| exp_acyclic_indvary2_step2_batch_yyy04 | 75.0070562 | 74.78125 | Memory |
| exp_acyclic_indvary2_step3_batch_yyy01 | 5.170090699 | 5.171875 | Memory |
| exp_acyclic_indvary2_step3_batch_yyy02 | 60.42786584 | 60.375 | Memory |
| exp_acyclic_indvary2_step3_batch_yyy03 | 67.36587816 | 67.296875 | Memory |
| exp_acyclic_indvary2_step3_batch_yyy04 | 64.88359852 | 64.8125 | Memory |
| exp_acyclic_indvary3_step1_batch_yyy01 | 0.097791303 | 0.109375 | |
| exp_acyclic_indvary3_step1_batch_yyy02 | 0.062898651 | 0.0625 | |
| exp_acyclic_indvary3_step1_batch_yyy03 | 0.07002136 | 0.078125 | |
| exp_acyclic_indvary3_step1_batch_yyy04 | 0.052853784 | 0.0625 | |
| exp_acyclic_indvary3_step2_batch_yyy01 | 0.315343464 | 0.3125 | |
| exp_acyclic_indvary3_step2_batch_yyy02 | 0.477033472 | 0.484375 | |
| exp_acyclic_indvary3_step2_batch_yyy03 | 20.27807138 | 20.1875 | |
| exp_acyclic_indvary3_step2_batch_yyy04 | 0.982323777 | 0.984375 | |
| exp_acyclic_indvary3_step3_batch_yyy01 | 24.2120256 | 24.21875 | |
| exp_acyclic_indvary3_step3_batch_yyy02 | 48.4700877 | 48.4375 | |
| exp_acyclic_indvary3_step3_batch_yyy03 | 47.0217335 | 47.03125 | |
| exp_acyclic_indvary3_step3_batch_yyy04 | 99.68006516 | 99.125 | |
| exp_cycles_depvary_step1_batch_yyy01 | 0.011494451 | 0.015625 | |
| exp_cycles_depvary_step1_batch_yyy02 | 0.013124669 | 0.015625 | |
| exp_cycles_depvary_step1_batch_yyy03 | 0.036522049 | 0.046875 | |
| exp_cycles_depvary_step1_batch_yyy04 | 0.012900108 | 0.015625 | |
| exp_cycles_depvary_step2_batch_yyy01 | 0.015194975 | 0.015625 | Runtime |
| exp_cycles_depvary_step2_batch_yyy02 | 0.047197491 | 0.046875 | |
| exp_cycles_depvary_step2_batch_yyy03 | 0.023367822 | 0.03125 | Runtime |
| exp_cycles_depvary_step2_batch_yyy04 | 0.080197597 | 0.078125 | |
| exp_cycles_depvary_step3_batch_yyy01 | 25.64668244 | 25.640625 | |
| exp_cycles_depvary_step3_batch_yyy02 | 52.71346553 | 52.6875 | Memory |
| exp_cycles_depvary_step3_batch_yyy03 | 10.34591035 | 10.34375 | |
| exp_cycles_depvary_step3_batch_yyy04 | 8.324258551 | 8.25 | |
| exp_cycles_indvary1_step1_batch_yyy01 | 8.914319354 | 8.796875 | Runtime |
| exp_cycles_indvary1_step1_batch_yyy02 | 0.095706628 | 0.09375 | Runtime |
| exp_cycles_indvary1_step1_batch_yyy03 | 11.03535573 | 11.03125 | Runtime |
| exp_cycles_indvary1_step1_batch_yyy04 | 9.239433041 | 9.21875 | |
| exp_cycles_indvary1_step2_batch_yyy01 | 6.558330059 | 6.5625 | Runtime |
| exp_cycles_indvary1_step2_batch_yyy02 | 52.05045095 | 52.046875 | Memory |
| exp_cycles_indvary1_step2_batch_yyy03 | 0.665531139 | 0.65625 | Runtime |
| exp_cycles_indvary1_step2_batch_yyy04 | 51.83307819 | 51.84375 | Memory |
| exp_cycles_indvary1_step3_batch_yyy01 | 48.11114094 | 48.09375 | Memory |
| exp_cycles_indvary1_step3_batch_yyy02 | 18.88910989 | 18.90625 | |
| exp_cycles_indvary1_step3_batch_yyy03 | 48.27523379 | 48.265625 | Memory |
| exp_cycles_indvary1_step3_batch_yyy04 | 31.48268279 | 31.4375 | Runtime |
| exp_cycles_indvary2_step1_batch_yyy01 | 52.69033418 | 52.640625 | Memory |
| exp_cycles_indvary2_step1_batch_yyy02 | 28.31878659 | 28.328125 | Runtime |
| exp_cycles_indvary2_step1_batch_yyy03 | 54.55117451 | 54.515625 | Memory |
| exp_cycles_indvary2_step1_batch_yyy04 | 7.880503524 | 7.875 | |
| exp_cycles_indvary2_step2_batch_yyy01 | 47.13945566 | 47.09375 | Memory |
| exp_cycles_indvary2_step2_batch_yyy02 | 43.99903607 | 43.9375 | Memory |
| exp_cycles_indvary2_step2_batch_yyy03 | 46.53313989 | 46.453125 | Memory |
| exp_cycles_indvary2_step2_batch_yyy04 | 43.49448267 | 43.421875 | Memory |
| exp_cycles_indvary2_step3_batch_yyy01 | 42.39867916 | 42.296875 | Memory |
| exp_cycles_indvary2_step3_batch_yyy02 | 43.04498391 | 43 | Memory |
| exp_cycles_indvary2_step3_batch_yyy03 | 42.51174344 | 42.421875 | Memory |
| exp_cycles_indvary2_step3_batch_yyy04 | 43.48903164 | 43.5 | Memory |
| exp_cycles_indvary3_step1_batch_yyy01 | 0.254123884 | 0.25 | |
| exp_cycles_indvary3_step1_batch_yyy02 | 7.87021357 | 7.796875 | |
| exp_cycles_indvary3_step1_batch_yyy03 | 0.194615807 | 0.1875 | |
| exp_cycles_indvary3_step1_batch_yyy04 | 2.588143214 | 2.59375 | |
| exp_cycles_indvary3_step2_batch_yyy01 | 70.25482712 | 70.21875 | Memory |
| exp_cycles_indvary3_step2_batch_yyy02 | 7.231557008 | 7.234375 | Memory |
| exp_cycles_indvary3_step2_batch_yyy03 | 70.63879081 | 70.5 | Memory |
| exp_cycles_indvary3_step2_batch_yyy04 | 63.33033769 | 63.328125 | Memory |
| exp_cycles_indvary3_step3_batch_yyy01 | 104.9427568 | 104.921875 | Memory |
| exp_cycles_indvary3_step3_batch_yyy02 | 85.08031172 | 84.984375 | Memory |
| exp_cycles_indvary3_step3_batch_yyy03 | 109.4636567 | 109.421875 | Memory |
| exp_cycles_indvary3_step3_batch_yyy04 | 119.5763803 | 119.46875 | Memory |

Aba-web (No pickle and cache)

| | Wall time (s) | CPU time (s) | Exception |
|---|---|---|---|
| exp_acyclic_depvary_step1_batch_yyy01 | 0.013839802 | 0 | |
| exp_acyclic_depvary_step1_batch_yyy02 | 0.013725264 | 0 | |
| exp_acyclic_depvary_step1_batch_yyy03 | 0.018488238 | 0.03125 | |
| exp_acyclic_depvary_step1_batch_yyy04 | 0.02204959 | 0.015625 | |
| exp_acyclic_depvary_step2_batch_yyy01 | 0.043877903 | 0.046875 | |
| exp_acyclic_depvary_step2_batch_yyy02 | 0.092252757 | 0.09375 | |
| exp_acyclic_depvary_step2_batch_yyy03 | 0.117446599 | 0.109375 | |
| exp_acyclic_depvary_step2_batch_yyy04 | 0.088802659 | 0.078125 | |
| exp_acyclic_depvary_step3_batch_yyy01 | 1.630324734 | 1.453125 | |
| exp_acyclic_depvary_step3_batch_yyy02 | 1.504347723 | 1.421875 | |
| exp_acyclic_depvary_step3_batch_yyy03 | 1.705245717 | 1.640625 | |
| exp_acyclic_depvary_step3_batch_yyy04 | 1.137852425 | 1.125 | |
| exp_acyclic_indvary1_step1_batch_yyy01 | 13.80269209 | 13.28125 | Runtime |
| exp_acyclic_indvary1_step1_batch_yyy02 | 7.424194529 | 7.375 | Runtime |
| exp_acyclic_indvary1_step1_batch_yyy03 | 17.90438993 | 17.8125 | |
| exp_acyclic_indvary1_step1_batch_yyy04 | 33.56632017 | 33.421875 | Runtime |
| exp_acyclic_indvary1_step2_batch_yyy01 | 272.329284 | 270.921875 | Memory & timeout |
| exp_acyclic_indvary1_step2_batch_yyy02 | 2122.368454 | 2114.46875 | Memory & timeout |
| exp_acyclic_indvary1_step2_batch_yyy03 | 914.3815474 | 872.96875 | Memory & timeout |
| exp_acyclic_indvary1_step2_batch_yyy04 | 330.050572 | 329.40625 | Runtime & timeout |
| exp_acyclic_indvary1_step3_batch_yyy01 | 12.18143516 | 11.921875 | |
| exp_acyclic_indvary1_step3_batch_yyy02 | 4.106505963 | 4.09375 | |
| exp_acyclic_indvary1_step3_batch_yyy03 | 52.07730658 | 52.03125 | |
| exp_acyclic_indvary1_step3_batch_yyy04 | 14.38501638 | 14.375 | Runtime |
| exp_acyclic_indvary2_step1_batch_yyy01 | 12.52045612 | 12.515625 | |
| exp_acyclic_indvary2_step1_batch_yyy02 | 1.637435942 | 1.640625 | |
| exp_acyclic_indvary2_step1_batch_yyy03 | 62.10012468 | 62.03125 | |
| exp_acyclic_indvary2_step1_batch_yyy04 | 11.92700851 | 11.90625 | |
| exp_acyclic_indvary2_step2_batch_yyy01 | 324.8199992 | 324.53125 | |
| exp_acyclic_indvary2_step2_batch_yyy02 | 375.0289687 | 374.875 | Memory & timeout |
| exp_acyclic_indvary2_step2_batch_yyy03 | 449.0996357 | 447.828125 | Memory & timeout |
| exp_acyclic_indvary2_step2_batch_yyy04 | 30.71486527 | 30.71875 | Memory |
| exp_acyclic_indvary2_step3_batch_yyy01 | 12.11011904 | 12.109375 | Memory |
| exp_acyclic_indvary2_step3_batch_yyy02 | 13.07575436 | 13.0625 | Memory |
| exp_acyclic_indvary2_step3_batch_yyy03 | 14.00717812 | 14.015625 | Memory |
| exp_acyclic_indvary2_step3_batch_yyy04 | 612.3102826 | 589.9375 | Memory & timeout |
| exp_acyclic_indvary3_step1_batch_yyy01 | 0.391439948 | 0.390625 | |
| exp_acyclic_indvary3_step1_batch_yyy02 | 0.27996164 | 0.28125 | |
| exp_acyclic_indvary3_step1_batch_yyy03 | 0.370393284 | 0.34375 | |
| exp_acyclic_indvary3_step1_batch_yyy04 | 0.176443928 | 0.15625 | |
| exp_acyclic_indvary3_step2_batch_yyy01 | 3.009582601 | 2.921875 | |
| exp_acyclic_indvary3_step2_batch_yyy02 | 3.5367669 | 3.46875 | |
| exp_acyclic_indvary3_step2_batch_yyy03 | 189.6481771 | 180.328125 | Timeout |
| exp_acyclic_indvary3_step2_batch_yyy04 | 11.97307741 | 11.234375 | |
| exp_acyclic_indvary3_step3_batch_yyy01 | 354.9011549 | 344.0625 | Timeout |
| exp_acyclic_indvary3_step3_batch_yyy02 | 562.9825295 | 555.703125 | Timeout |
| exp_acyclic_indvary3_step3_batch_yyy03 | 514.2050107 | 509.703125 | Timeout |
| exp_acyclic_indvary3_step3_batch_yyy04 | 950.6512335 | 940.984375 | Timeout |
| exp_cycles_depvary_step1_batch_yyy01 | 0.027032608 | 0.03125 | |
| exp_cycles_depvary_step1_batch_yyy02 | 0.02655434 | 0.03125 | |
| exp_cycles_depvary_step1_batch_yyy03 | 0.106482973 | 0.109375 | |
| exp_cycles_depvary_step1_batch_yyy04 | 0.025445087 | 0.015625 | |
| exp_cycles_depvary_step2_batch_yyy01 | 0.092876353 | 0.09375 | Runtime |
| exp_cycles_depvary_step2_batch_yyy02 | 0.172957293 | 0.171875 | |
| exp_cycles_depvary_step2_batch_yyy03 | 0.061566018 | 0.0625 | Runtime |
| exp_cycles_depvary_step2_batch_yyy04 | 0.435894646 | 0.4375 | |
| exp_cycles_depvary_step3_batch_yyy01 | 175.4016851 | 172.59375 | Timeout |
| exp_cycles_depvary_step3_batch_yyy02 | 421.075742 | 412.609375 | Memory & timeout |
| exp_cycles_depvary_step3_batch_yyy03 | 81.54591147 | 80.40625 | |
| exp_cycles_depvary_step3_batch_yyy04 | 58.2971953 | 57.796875 | |
| exp_cycles_indvary1_step1_batch_yyy01 | 20.0047757 | 19.859375 | Runtime |
| exp_cycles_indvary1_step1_batch_yyy02 | 0.333843349 | 0.328125 | Runtime |
| exp_cycles_indvary1_step1_batch_yyy03 | 18.73090469 | 18.546875 | Runtime |
| exp_cycles_indvary1_step1_batch_yyy04 | 13.45408566 | 13.390625 | |
| exp_cycles_indvary1_step2_batch_yyy01 | 111.018808 | 110.328125 | Runtime |
| exp_cycles_indvary1_step2_batch_yyy02 | 390.6238527 | 387.234375 | Memory & timeout |
| exp_cycles_indvary1_step2_batch_yyy03 | 43.48690182 | 41.890625 | Runtime |
| exp_cycles_indvary1_step2_batch_yyy04 | 416.0816036 | 411.296875 | Memory & timeout |
| exp_cycles_indvary1_step3_batch_yyy01 | 377.5164493 | 376.640625 | Memory & timeout |
| exp_cycles_indvary1_step3_batch_yyy02 | 11.65868951 | 11.65625 | Runtime |
| exp_cycles_indvary1_step3_batch_yyy03 | 450.0705616 | 423.03125 | Memory & timeout |
| exp_cycles_indvary1_step3_batch_yyy04 | 366.2375796 | 360.9375 | Runtime & timeout |
| exp_cycles_indvary2_step1_batch_yyy01 | 30.8657915 | 30.4375 | Memory |
| exp_cycles_indvary2_step1_batch_yyy02 | 550.327317 | 538.09375 | Memory & timeout |
| exp_cycles_indvary2_step1_batch_yyy03 | 532.485801 | 515.9375 | Memory & timeout |
| exp_cycles_indvary2_step1_batch_yyy04 | 46.49705239 | 44.53125 | Runtime |
| exp_cycles_indvary2_step2_batch_yyy01 | 388.0164577 | 379.328125 | Memory & timeout |
| exp_cycles_indvary2_step2_batch_yyy02 | 393.050634 | 384.40625 | Memory & timeout |
| exp_cycles_indvary2_step2_batch_yyy03 | 428.6728049 | 416.359375 | Memory & timeout |
| exp_cycles_indvary2_step2_batch_yyy04 | 446.2884895 | 433.609375 | Memory & timeout |
| exp_cycles_indvary2_step3_batch_yyy01 | 589.34546 | 556.859375 | Memory & timeout |
| exp_cycles_indvary2_step3_batch_yyy02 | 404.4377012 | 391.921875 | Memory & timeout |
| exp_cycles_indvary2_step3_batch_yyy03 | 397.7047183 | 383.796875 | Memory & timeout |
| exp_cycles_indvary2_step3_batch_yyy04 | 361.1104872 | 355.765625 | Memory & timeout |
| exp_cycles_indvary3_step1_batch_yyy01 | 0.910930533 | 0.90625 | |
| exp_cycles_indvary3_step1_batch_yyy02 | 45.81873241 | 45.0625 | |
| exp_cycles_indvary3_step1_batch_yyy03 | 0.788313763 | 0.78125 | |
| exp_cycles_indvary3_step1_batch_yyy04 | 12.05009947 | 11.75 | Runtime |
| exp_cycles_indvary3_step2_batch_yyy01 | 197.4267437 | 192.15625 | Runtime & timeout |
| exp_cycles_indvary3_step2_batch_yyy02 | 467.8784945 | 461.0625 | Memory & timeout |
| exp_cycles_indvary3_step2_batch_yyy03 | 476.7613046 | 475.78125 | Memory & timeout |
| exp_cycles_indvary3_step2_batch_yyy04 | 463.1522997 | 462.140625 | Memory & timeout |
| exp_cycles_indvary3_step3_batch_yyy01 | 656.3492615 | 649.828125 | Memory & timeout |
| exp_cycles_indvary3_step3_batch_yyy02 | 531.8614243 | 531.046875 | Memory & timeout |
| exp_cycles_indvary3_step3_batch_yyy03 | 515.3057859 | 514.828125 | Memory & timeout |
| exp_cycles_indvary3_step3_batch_yyy04 | 596.9302604 | 596.140625 | Memory & timeout |

Aba-web (Pickle and no cache)

| | Wall time (s) | CPU time (s) | Exception |
|---|---|---|---|
| exp_acyclic_depvary_step1_batch_yyy01 | 0.01319609 | 0.015625 | |
| exp_acyclic_depvary_step1_batch_yyy02 | 0.0099685 | 0.015625 | |
| exp_acyclic_depvary_step1_batch_yyy03 | 0.010903278 | 0 | |
| exp_acyclic_depvary_step1_batch_yyy04 | 0.01190374 | 0 | |
| exp_acyclic_depvary_step2_batch_yyy01 | 0.024984882 | 0.015625 | |
| exp_acyclic_depvary_step2_batch_yyy02 | 0.039461827 | 0.046875 | |
| exp_acyclic_depvary_step2_batch_yyy03 | 0.07115724 | 0.078125 | |
| exp_acyclic_depvary_step2_batch_yyy04 | 0.023005715 | 0.015625 | |
| exp_acyclic_depvary_step3_batch_yyy01 | 0.160948046 | 0.15625 | |
| exp_acyclic_depvary_step3_batch_yyy02 | 0.165098509 | 0.171875 | |
| exp_acyclic_depvary_step3_batch_yyy03 | 0.214614637 | 0.203125 | |
| exp_acyclic_depvary_step3_batch_yyy04 | 0.191707039 | 0.203125 | |
| exp_acyclic_indvary1_step1_batch_yyy01 | 6.449963524 | 6.4375 | runtime |
| exp_acyclic_indvary1_step1_batch_yyy02 | 3.197325558 | 3.203125 | runtime |
| exp_acyclic_indvary1_step1_batch_yyy03 | 14.95068667 | 14.921875 | |
| exp_acyclic_indvary1_step1_batch_yyy04 | 7.222392957 | 7.21875 | runtime |
| exp_acyclic_indvary1_step2_batch_yyy01 | 149.4031561 | 149.0625 | memory & timeout |
| exp_acyclic_indvary1_step2_batch_yyy02 | 1853.777154 | 1833.421875 | memory & timeout |
| exp_acyclic_indvary1_step2_batch_yyy03 | 285.873713 | 282.796875 | memory & timeout |
| exp_acyclic_indvary1_step2_batch_yyy04 | 3380.536731 | 3350.703125 | runtime & timeout |
| exp_acyclic_indvary1_step3_batch_yyy01 | 1.801212953 | 1.796875 | |
| exp_acyclic_indvary1_step3_batch_yyy02 | 0.444739935 | 0.4375 | runtime |
| exp_acyclic_indvary1_step3_batch_yyy03 | 1.133093146 | 1.125 | runtime |
| exp_acyclic_indvary1_step3_batch_yyy04 | 0.233652165 | 0.234375 | runtime |
| exp_acyclic_indvary2_step1_batch_yyy01 | 0.254460311 | 0.25 | runtime |
| exp_acyclic_indvary2_step1_batch_yyy02 | 0.337710135 | 0.34375 | |
| exp_acyclic_indvary2_step1_batch_yyy03 | 6.655553882 | 6.578125 | runtime |
| exp_acyclic_indvary2_step1_batch_yyy04 | 0.151024704 | 0.140625 | runtime |
| exp_acyclic_indvary2_step2_batch_yyy01 | 0.153349949 | 0.140625 | runtime |
| exp_acyclic_indvary2_step2_batch_yyy02 | 109.6607554 | 104.78125 | memory |
| exp_acyclic_indvary2_step2_batch_yyy03 | 93.16198087 | 86.984375 | memory |
| exp_acyclic_indvary2_step2_batch_yyy04 | 87.64037179 | 85.71875 | memory |
| exp_acyclic_indvary2_step3_batch_yyy01 | 78.6211754 | 77.890625 | memory |
| exp_acyclic_indvary2_step3_batch_yyy02 | 76.99444429 | 74.5625 | memory |
| exp_acyclic_indvary2_step3_batch_yyy03 | 67.36950803 | 67.03125 | memory |
| exp_acyclic_indvary2_step3_batch_yyy04 | 62.42616777 | 62.140625 | memory |
| exp_acyclic_indvary3_step1_batch_yyy01 | 0.09431276 | 0.078125 | |
| exp_acyclic_indvary3_step1_batch_yyy02 | 0.070244192 | 0.078125 | |
| exp_acyclic_indvary3_step1_batch_yyy03 | 0.073528844 | 0.0625 | |
| exp_acyclic_indvary3_step1_batch_yyy04 | 0.061393159 | 0.0625 | |
| exp_acyclic_indvary3_step2_batch_yyy01 | 0.299489218 | 0.296875 | |
| exp_acyclic_indvary3_step2_batch_yyy02 | 0.409904207 | 0.40625 | |
| exp_acyclic_indvary3_step2_batch_yyy03 | 18.69383269 | 18.515625 | |
| exp_acyclic_indvary3_step2_batch_yyy04 | 0.959979047 | 0.953125 | |
| exp_acyclic_indvary3_step3_batch_yyy01 | 26.19981083 | 26.078125 | |
| exp_acyclic_indvary3_step3_batch_yyy02 | 55.32143699 | 55.03125 | |
| exp_acyclic_indvary3_step3_batch_yyy03 | 48.28090328 | 48 | |
| exp_acyclic_indvary3_step3_batch_yyy04 | 94.88177878 | 94.46875 | |
| exp_cycles_depvary_step1_batch_yyy01 | 0.011142612 | 0.015625 | |
| exp_cycles_depvary_step1_batch_yyy02 | 0.013251917 | 0.015625 | |
| exp_cycles_depvary_step1_batch_yyy03 | 0.055356312 | 0.0625 | |
| exp_cycles_depvary_step1_batch_yyy04 | 0.012901735 | 0.015625 | |
| exp_cycles_depvary_step2_batch_yyy01 | 0.026800237 | 0.03125 | runtime |
| exp_cycles_depvary_step2_batch_yyy02 | 0.036835649 | 0.03125 | runtime |
| exp_cycles_depvary_step2_batch_yyy03 | 0.0251302 | 0.015625 | runtime |
| exp_cycles_depvary_step2_batch_yyy04 | 0.08595108 | 0.09375 | |
| exp_cycles_depvary_step3_batch_yyy01 | 6.345723137 | 6.296875 | memory |
| exp_cycles_depvary_step3_batch_yyy02 | 0.609710108 | 0.609375 | memory |
| exp_cycles_depvary_step3_batch_yyy03 | 0.861785133 | 0.84375 | memory |
| exp_cycles_depvary_step3_batch_yyy04 | 14.57454326 | 14.5 | |
| exp_cycles_indvary1_step1_batch_yyy01 | 5.913476641 | 5.875 | |
| exp_cycles_indvary1_step1_batch_yyy02 | 0.074881131 | 0.078125 | runtime |
| exp_cycles_indvary1_step1_batch_yyy03 | 11.52084632 | 11.421875 | runtime |
| exp_cycles_indvary1_step1_batch_yyy04 | 7.467165789 | 7.4375 | runtime |
| exp_cycles_indvary1_step2_batch_yyy01 | 8.984435566 | 8.9375 | runtime |
| exp_cycles_indvary1_step2_batch_yyy02 | 63.82037248 | 63.296875 | memory |
| exp_cycles_indvary1_step2_batch_yyy03 | 4.923920531 | 4.921875 | runtime |
| exp_cycles_indvary1_step2_batch_yyy04 | 55.04461273 | 54.765625 | memory |
| exp_cycles_indvary1_step3_batch_yyy01 | 51.7399106 | 51.40625 | memory |
| exp_cycles_indvary1_step3_batch_yyy02 | 26.91493197 | 26.78125 | |
| exp_cycles_indvary1_step3_batch_yyy03 | 47.22915553 | 47 | memory |
| exp_cycles_indvary1_step3_batch_yyy04 | 49.89344482 | 49.421875 | memory |
| exp_cycles_indvary2_step1_batch_yyy01 | 47.87894284 | 47.734375 | memory |
| exp_cycles_indvary2_step1_batch_yyy02 | 26.95778558 | 26.734375 | runtime |
| exp_cycles_indvary2_step1_batch_yyy03 | 53.04542148 | 52.75 | memory |
| exp_cycles_indvary2_step1_batch_yyy04 | 7.953339951 | 7.90625 | |
| exp_cycles_indvary2_step2_batch_yyy01 | 45.44846039 | 45.15625 | memory |
| exp_cycles_indvary2_step2_batch_yyy02 | 45.42555608 | 45.25 | memory |
| exp_cycles_indvary2_step2_batch_yyy03 | 42.637643 | 42.453125 | memory |
| exp_cycles_indvary2_step2_batch_yyy04 | 44.78569869 | 44.609375 | memory |
| exp_cycles_indvary2_step3_batch_yyy01 | 45.72322501 | 45.5625 | memory |
| exp_cycles_indvary2_step3_batch_yyy02 | 45.00073074 | 44.859375 | memory |
| exp_cycles_indvary2_step3_batch_yyy03 | 44.89937246 | 44.71875 | memory |
| exp_cycles_indvary2_step3_batch_yyy04 | 44.40590031 | 44.203125 | memory |
| exp_cycles_indvary3_step1_batch_yyy01 | 0.209175023 | 0.21875 | |
| exp_cycles_indvary3_step1_batch_yyy02 | 7.024880597 | 6.984375 | |
| exp_cycles_indvary3_step1_batch_yyy03 | 0.203629991 | 0.203125 | |
| exp_cycles_indvary3_step1_batch_yyy04 | 1.622971467 | 1.59375 | |
| exp_cycles_indvary3_step2_batch_yyy01 | 57.07060546 | 56.734375 | runtime |
| exp_cycles_indvary3_step2_batch_yyy02 | 3.187962514 | 3.1875 | memory |
| exp_cycles_indvary3_step2_batch_yyy03 | 66.60766608 | 66.3125 | memory |
| exp_cycles_indvary3_step2_batch_yyy04 | 67.3013091 | 66.40625 | memory |
| exp_cycles_indvary3_step3_batch_yyy01 | 128.0351968 | 126.203125 | memory & timeout |
| exp_cycles_indvary3_step3_batch_yyy02 | 106.0845602 | 104.828125 | memory |
| exp_cycles_indvary3_step3_batch_yyy03 | 110.5838849 | 109.828125 | memory |
| exp_cycles_indvary3_step3_batch_yyy04 | 111.7377426 | 110.875 | memory |